\documentclass[10pt,compsoc]{IEEEtran}
\IEEEoverridecommandlockouts
\newcommand{\ApproxSign}{\raise.17ex\hbox{$\scriptstyle\sim$}}

\usepackage{cite}
\usepackage{multirow}
\usepackage{amsmath}
\usepackage{amssymb}
\usepackage{amsfonts}
\usepackage{algorithmic}
\usepackage{graphicx}
\usepackage{textcomp}
\usepackage{float}
\usepackage{blindtext}
\usepackage{xcolor}
\usepackage{caption}
\usepackage{array}
\usepackage{enumitem}
\usepackage{subcaption}
\usepackage{booktabs}
\newcolumntype{C}[1]{>{\centering\let\newline\\\arraybackslash\hspace{0pt}}m{#1}}
\newcolumntype{Q}{>{\columncolor{blue}}C}

\newcommand\blfootnote[1]{%
  \begingroup
  \renewcommand\thefootnote{}\footnote{#1}%
  \addtocounter{footnote}{-1}%
  \endgroup
}
\begin{document}

\title{Modeling Data Reuse in Deep Neural Networks  by Taking Data-Types into Cognizance \\
{}
}
\author{\IEEEauthorblockN{Nandan Kumar Jha$^\star$, Sparsh Mittal$^\dagger$} \\
\IEEEauthorblockA{
$^\star$Indian Institute of Technology Hyderabad, India,
$^\dagger$Indian Institute of Technology Roorkee, India \\
Email: cs17mtech11010@iith.ac.in,sparshfec@iitr.ac.in
}
}

\IEEEtitleabstractindextext{
\begin{abstract}
In recent years, researchers have focused on reducing the model size and number of computations (measured as ``multiply-accumulate''  or MAC operations) of DNNs. The energy consumption of a DNN depends on both the number of MAC operations and the energy efficiency of each MAC operation. The former can be estimated at design time; however, the latter depends on the intricate data reuse patterns and underlying hardware architecture. Hence, estimating it at design time is challenging.  This work shows that the conventional approach to estimate the data reuse, 
viz. arithmetic intensity, does not always correctly estimate the degree of data reuse in DNNs since it gives equal importance to all the data types. We propose a novel model, termed ``data type aware weighted arithmetic intensity'' ($DI$), which accounts for the unequal importance of different data types in DNNs. We evaluate our model on 25 state-of-the-art DNNs on two GPUs. We show that our model accurately models data-reuse for all possible data reuse patterns for different types of convolution and different types of layers. We show that our model is a better indicator of the energy efficiency of DNNs. We also show its generality using the central limit theorem. 

\end{abstract}

 }

\maketitle
%\IEEEdisplaynontitleabstractindextext

\section{Introduction}
DNNs are now being used in a wide range of cognitive applications\blfootnote{This work was supported by  Semiconductor Research Corporation.}. After the success of AlexNet \cite{NIPS2012_4824},  the research in DNNs has focused on achieving higher accuracy even at the cost of large DNN size and computational complexity. The focus on accuracy has led to over-parameterized DNNs, e.g., VGG-16 
\cite{Simonyan2014VeryDC}, 
Inception-v4 \cite{Szegedy2017Inceptionv4IA}, ResNet152-v2 
\cite{He2016IdentityMI} etc. By contrast, 
recent networks such as SqueezeNet \cite{Iandola2016SqueezeNetAA}, MobileNet-V1 
\cite{Howard2017MobileNetsEC}, MobileNet-V2 \cite{8578572}, ShuffleNets \cite{Ma_2018_ECCV}, etc. 
 focus on making the DNN compact by reducing the number of parameters, or MACs or both. However, reducing the number of MACs does not 
necessarily  
make DNNs energy efficient because energy is dominated by data movement rather than computation \cite{horowitz2014computing}. The data movement primarily depends on the degree of data reuse present in the workloads. 

To enable the deployment of DNN models in a wide range of applications such as autonomous driving and drones, the energy consumption of DNN inference must be within a prescribed envelope. Hence, DNNs need to be carefully examined based on the number of computations (i.e., MACs) and the energy efficiency of MAC operations. Unfortunately, the latter metric has mainly been overlooked in DNN design because a study of energy efficiency requires precise knowledge of the degree of data reuse and parallelism present in the DNNs, and how the underlying hardware platform exploits this parallelism. Further, the implications of reducing the number of parameters and MACs on the energy efficiency of DNN is not well-understood.

Traditionally, arithmetic intensity \cite{harris2005GPU} is used to model the degree of data reuse. It is also used in the ``roofline model'' \cite{Williams:2009:RIV:1498765.1498785} for predicting whether a workload is compute-bound or memory-bound. Therefore, it represents the {\em degree of data reuse} available in workloads and hence {\em bandwidth pressure}. Lower arithmetic intensity implies a lower degree of data reuse and high bandwidth pressure and vice versa. The arithmetic intensity considers the memory footprint and the number of arithmetic operations and shows the degree of data reuse available in a workload. In other words, arithmetic intensity shows how efficiently arithmetic operation can reuse the data fetched from different levels in the memory hierarchy. However, the memory footprint does not always reflect the actual number of off-chip accesses, which largely depends on the data reuse available in workload and how well the underlying platform exploits the data reuse available in the workload. Arithmetic intensity can represent power/energy efficiency only when all the data types have the same access behavior. For example, Choi et al. \cite{6569852} and Ghane et al. \cite{DBLP:journals/corr/abs-1809-09206} use arithmetic intensity to model the power/energy efficiency.

DNNs have different types of data such as filter weights, input and output activations, partial sums, which have different reuse patterns \cite{Yang2017AMT} and hence, reuse importance. Also, the layers in DNNs have distinct computation and reuse patterns with different bandwidth requirements. For example, convolution (Conv) layers have a high degree of reuse, and they are compute-bound, whereas fully connected (FC) layers have low-reuse and a high number of parameters, and hence, they are memory-bound 
\cite{park2018deep}. Moreover, the same layer with different types of convolution possesses a different degree of data reuse (refer Section \ref{sec:background} for more details).
Further, due to the different DNN topologies such as branching, skip connections, and dense connections \cite{mlsys2020_15}, the number of concurrent activations varies during the runtime \cite{DBLP:journals/corr/abs-1801-04326} which in turn, affects the reuse behavior of DNNs. Given these factors, {\em it is interesting to investigate whether arithmetic intensity can be used as a representative of energy efficiency in DNN models}, or, is there a need for a more accurate metric.

\begin{figure}[htbp]
\centering
\includegraphics[scale=0.33]{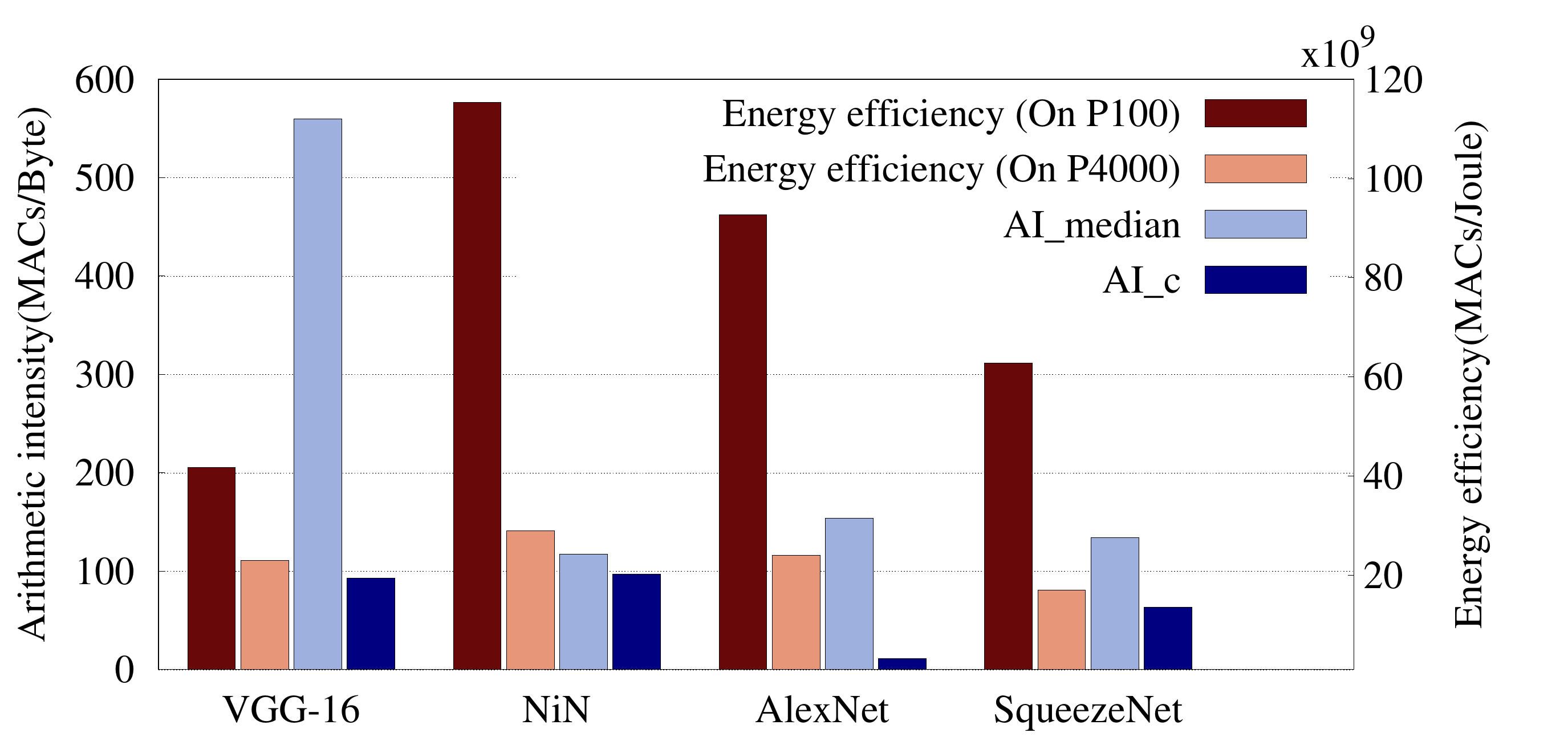}
\caption{ Neither $AI_{median}$ nor $AI_c$ is  a representative of data reuse in DNNs}
\label{fig:AIexamples}
\end{figure}

In Figure \ref{fig:AIexamples}, cumulative arithmetic intensity ($AI_c$) and 
median value of layer-wise arithmetic intensity ($AI_{median}$) of DNNs are 
shown (these metrics are explained in Section \ref{sec:PreMethodology}). VGG-16 \cite{Simonyan2014VeryDC} and NiN 
\cite{DBLP:journals/corr/LinCY13} have almost equal $AI_c$ but NiN has significantly higher energy efficiency (measured on both GPU P100 and P4000) than VGG-16, even when VGG-16 has quite higher $AI_{median}$.  Also, AlexNet has lower $AI_c$ and  $AI_{median}$ than VGG-16 but has higher energy efficiency than VGG-16. It shows that both layer-wise arithmetic intensity and cumulative arithmetic intensity are not good indicators of energy efficiency of 
DNNs and hence, {\em there is a need for better model/metric to estimate the data reuse in DNNs to understand their energy efficiency. }

Our main {\bf contributions} can be summarized as follows. 
\begin{enumerate}
\item We first explore the intricacies of data reuse patterns   (Section \ref{sec:background}) and energy-efficiency of various DNNs (Section \ref{sec:EnergyEfficiency}). We also perform a detailed analysis of layer-wise data reuse patterns in DNNs (Section \ref{sec:PreMethodology}).  We include all the possible variations of data reuse patterns arising from (a) different types of convolutions such as standard, group, pointwise, depthwise, etc., (b) different types of layers such as Conv, FC, and others, and (c) different design heuristics such as feed-forward/skip connections. Our comprehensive experiment and analysis show that {\em data reuse estimated by arithmetic intensity is not tightly-coupled with the energy efficiency of MAC operations in DNNs}. 
\item  We experimentally measure the energy consumption and energy efficiency of DNNs on two GPUs, P100 and P4000, and one CPU. We show that {\em activation reuse has a higher impact on energy efficiency than weight reuse}.
\item We propose a novel metric, termed ``data type aware weighted arithmetic intensity'' ($DI$), which takes data types into cognizance and accounts for the unequal importance of different data types in DNNs. Our proposed metric ($DI$) more accurately quantifies the intrinsic relationship between data reuse and energy efficiency of MAC operations (Section \ref{sec:methodology}). 
\item We validate our model on 25 state-of-the-art DNNs, including both highly-accurate DNNs and compact DNNs, which have between 221 to 15,470 million weights and between 0.54 to 138 million MACs.  
\item We also use the ``central limit theorem'' to prove the generality of our proposed model (Section \ref{sec:GeneralityTesting}).
\end{enumerate}

\section{Background and motivation } \label{sec:background}
Table \ref{tab:SymbolNew} lists the symbols used. For simplicity and ease of comparison, we assume that a) height and width of filter are same, b) height and width of output feature map (ofmap) are same, and c) spatial size of input feature map (ifmap) and ofmap are equal.

%\small
\begin{table} [htbp] 
\caption{Symbols (Fmap = feature map, Ops = operations)}
\label{tab:SymbolNew}\centering 
\resizebox{0.46\textwidth}{!}{
\begin{tabular}{ |C{3.6cm}|C{1.9cm}|C{3.6cm}|C{1.8cm}| } \hline
 \textbf{Quantity (symbol)} & \textbf{Unit} & \textbf{Quantity (symbol)} & \textbf{Unit}  \\ \hline 
 Energy per pixel (EPP)   &    Joule &  Energy efficiency    &    MACs/Joule        \\ \hline 
 Average power ($P_{avg}$) & watt & Inference time ($I_t$) & Millisecond \\ \hline              
 \# Weights  ($W$) & Millions  & Throughput        &      MACs/sec \\ \hline
 \# MACs ($M_c$)   &  Millions & \# Activations ($A$) &Millions \\ \hline
 Fmap height     ($S_o$) &   -     & Fmap width     ($S_o$)   &  -    \\\hline
 Filter height ($S_k$) &  - & Filter width ($S_k$)   &   -  \\\hline
  \# filter-channels  ($M$)   &    -  & \# filters ($N$)   &   -    \\ \hline
  Compute to memory ratio (CMR) & OPs/sec/Byte & Group Size  ($G$)   &  - \\ \hline 
  Cumulative arithmetic intensity ($AI_c$) & MACs/Byte & Disparity factor ($d_f$) & -   \\ \hline
 Pearson Product Moment correlation coefficient ($r_p$) & - & Spearman's Rank correlation coefficient ($r_s$) & - \\\hline  
 
\end{tabular} }
\end{table}

\subsection{Data reuse patterns in DNNs}\label{sec:CONVtypeLayers}  
\textbf{Types of convolution:} 
Broadly, there are four types of convolutions. They are discussed below and  their properties are summarized in Table \ref{tab:ConvType}. Here, weight (learnable filter coefficients) reuse and activation (ifmaps and ofmaps) reuse are estimated as $M_c$/$W$ and  $M_c$/$A$, respectively. We use arithmetic intensity defined as $\frac{M_c}{W+A}$, as a metric to evaluate the bandwidth requirement of MACs \cite{harris2005GPU}.

\textbf{1. Standard (spatial) convolution:} In standard convolution, filtering (i.e., feature extraction) and combining (i.e., feature aggregation) are performed together. The total number of filter weights and activations (combined ifmaps and ofmaps) involved in standard convolution is $M\times N\times S_k^2$ and $(M + N)\times S_{o}^2$, respectively. The total number of MAC operations in standard convolution is  $M\times N\times S_k^2\times S_{o}^2$. Because of combined feature extraction and aggregation, standard convolution incurs high computational complexity.

\textbf{2. Pointwise convolution (PWConv):} In PWConv, the smaller receptive size of filter reduces the number of MACs as well as the number of filter weights involved,  which are $M\times N\times S_{o}^2$ and $M\times N$, respectively. The number of activations is the same as that in standard convolution. Pointwise convolution has been used in NiN \cite{DBLP:journals/corr/LinCY13}, inception modules in \cite{Szegedy_2015_CVPR,Szegedy_2016_CVPR,Szegedy2017Inceptionv4IA} and in SqueezeNet \cite{Iandola2016SqueezeNetAA}.

\textbf{3. Group convolution:} In group convolution, each group of 2-D filters convolve with only one group ($G$ number) of input feature maps. Compared to standard convolution, in group convolution, the number of MACs and filter weights are reduced by a factor of $g$ ($g$=$\frac{N}{G}$). The number of activations remains same as that in standard convolution. Group convolution has been used in AlexNet \cite{NIPS2012_4824}  ($g = 2$), ResNext ($g$ = 32) \cite{Xie2017AggregatedRT} and 1.0-G-SqNxt-23 ($g$ = 2) \cite{Gholami2018SqueezeNextHN}.

\textbf{4. Depthwise convolution:} Depthwise convolution (DWConv) performs only feature extraction where, one filter convolves with only one input feature map, i.e., one channel of input. Compared to standard convolution, it reduces the number of MACs and number of weights by a factor of $N$ (Table \ref{tab:ConvType}). Total number of activations involved in DWConv  is $2\times M\times S_{o}^2$. Depthwise convolution has been used in MobileNet-V1 \cite{Howard2017MobileNetsEC},  MobileNet-V2 \cite{8578572}, and XceptionNet \cite{Chollet_2017_CVPR}. Note that, DWConv is an extreme case of group convolution where $G$ = 1, i.e. $g$ = $N$ = $M$. 
\begin{table} [htbp]
\caption{Data reuse characteristics of convolution}
\label{tab:ConvType}
\centering 
\begin{tabular}{ |c| c| p {0.25 cm}| c| } 
 \hline
  Convolution &  Arithmetic intensity & $\frac{M_c}{W}$ & $\frac{M_c}{A}$  \\
 \hline
 Standard &  $\frac{M\times N\times S_k^2\times S_{o}^2}{M\times N\times S_k^2  +  (M+N)\times S_{o}^2}$ & $S_{o}^2$&
 $\big(\frac{M\times N}{M+N}\big) \times S_k^2$ \\                
 \hline
 Pointwise       &  $\frac{M\times N\times S_{o}^2}{ M\times N  +  (M+N)\times S_{o}^2 }$ & $S_{o}^2$ & $\big(\frac{M\times N}{M+N}\big) $    \\
 \hline
 Group            &  $\frac{M\times N\times S_k^2\times S_{o}^2}{ M\times N\times S_k^2  +  g\times(M+N)\times S_{o}^2 }$            & $S_{o}^2$ &$\big(\frac{M\times N}{M+N}\big) \times {\frac{S_k^2}{g}} $  \\
 \hline
 Depthwise       & $\frac{M\times S_k^2\times S_{o}^2}{ M\times S_k^2  +  (M+M)\times S_{o}^2 }$ & $S_{o}^2$ & $\big(\frac{M}{M+M}\big) \times S_k^2$  \\
 \hline
\end{tabular} 

\end{table}

{\bf Others layers:} The non-Conv layers such as pooling, ReLU, BatchNorm have a negligible number of learnable parameters, and there is no MAC operation involved. However, there are other operations, such as element-wise addition, comparison, and division. These layers have a low arithmetic intensity and high bandwidth requirement. FC layers have a very high number of parameters (weights), and fewer activations, making them memory bound. Hence, in FC layer, both the arithmetic intensity and weight reuse are approximately equal to 1 (as $M_c \approx W$ and $A \ll W$).

\subsection{Motivation}
Different types of convolutions have been applied to accomplish different design goals and achieve a trade-off between performance and computation/bandwidth overhead. Apart from this, various design heuristics have been used, such as residual connections \cite{He_2016_CVPR,He2016IdentityMI} to facilitate backpropagation in deeper networks, dense connections \cite{8099726} to enable feature reuse, etc. These design heuristics lead to different computational complexity and degrees of data reuse. Even in standard convolution, the degree of data reuse depends on multiple factors such as filter's dimensions, convolution stride, and dimensions of ifmaps. For the comparison of data reuse and computational complexity in different types of convolution, we assume specific values of variables and show the values of metrics normalized to that for standard convolution in Table \ref{tab:RelativeValueOfConvType}.

\begin{table} [htbp]
\caption{Metrics values normalized to  standard convolution (assuming $N$ = $M$ = 256, feature map size ($S_o\times S_o$) = $28\times 28$, filter size ($S_k\times S_k$) = $3\times 3$ and group size ($g$) = 4)}
\label{tab:RelativeValueOfConvType}
\centering 
\begin{tabular}{ |c| c| c| c| c | } 
 \hline
 \textbf{Convolution} & $\textbf{Arithmetic intensity}$ & $\textbf{$M_c$}$  & $\frac{M_c}{W}$ & $\frac{M_c}{A}$ \\
 \hline
 Standard          &1.00 &1.000   &1 &1.000 \\                
 %\hline
 Pointwise        &0.24 &0.111  &1 &0.111           \\
 %\hline
 Group           &0.45 &0.250  &1 &0.250  \\
 %\hline
 Depthwise        &0.01 &0.004  &1 &0.004   \\
 \hline
\end{tabular}
\end{table}

Evidently, the relative number of MACs decreases from standard convolution to DWConv, but the arithmetic intensity also reduces, which increases bandwidth requirement. It is well-known that energy is dominated by data movement rather than computation \cite{horowitz2014computing}. Hence, {\em a decrease in the number of MACs can be dwarfed by the increase in memory accesses, increasing the overall energy consumption}.  These observations motivate us to investigate a model that can better incorporate the dynamics of data reuse in DNNs and is a better indicator of the energy efficiency of MACs in DNNs.

\subsection{Experimental setup and metrics}\label{sec:setup}

We perform our experiments using  Caffe \cite{Jia:2014:CCA:2647868.2654889} on two GPUs, viz., Tesla P100  and Quadro P4000 which have significantly different compute and memory resources, as shown in  Table \ref{tab:GPUsFeatures}. Former is a data-center scale GPU while latter is a desktop GPU. We also validate our model on CPU in Section \ref{sec:ApplicabilityOfModel}.

\begin{table} [htbp]
  \caption{Configuration of GPUs used in our experiments}
   \label{tab:GPUsFeatures} 
   \resizebox{0.48\textwidth}{!}{
  \begin{tabular}{ |c|c|c|c|c|c| } 
    \hline
    
 \textbf{GPU}   & \textbf{\# core} & \textbf{L2 size} & \textbf{Peak bandwidth }  & \textbf{Peak Throughput} & \textbf{CMR }   \\
      \hline
    P4000 & 1792 & 2 MB & 243 GB/s & 5.2 TFLOPS & 21.4 FLOPs/Byte   \\ 
    \hline
     P100 & 3584 & 4 MB & 549 GB/s & 9.3 TFLOPS & 16.94 FLOPs/Byte   \\  
    \hline
  \end{tabular}}
\end{table}

{\bf Power and inference time readings:}
GPUs have massive compute and memory resources (Table \ref{tab:GPUsFeatures}); hence, smaller batch sizes can result in resource underutilization, which can lead to an unfair comparison of energy consumption in DNNs with different model size and memory-footprint. In general, larger models have better resource utilization than compact models at smaller $B$. Hence, for better utilization of GPU compute resources and to enable fair comparison, we take input batch size of four (as used in \cite{Ma_2018_ECCV}). For power and inference time measurement, we use {\tt nvidia-smi} utility, which is a high-level utility. The sampling rate of {\tt nvidia-smi} utility depends on the sampling rate of inbuilt power sensors in high-end GPUs, which is quite low \cite{bridges2016understanding}. For instance, the sampling rate on P100 GPU is $\approx50Hz$ (119 samples in 2.36 second), and on P4000 GPU, it is $\approx1Hz$ (119 samples in 118.23 sec). Therefore, as suggested in \cite{kasichayanula2012power}, we run the DNN computations for a longer time to see the changes in power reading. In particular, we run 500 iterations for DNNs with very low inference time (AlexNet, NiN, SqueezeNet variants) and 200 iterations for the rest of DNNs shown in Table \ref{tab:DataReuse}.

For power and inference time readings, we used a similar methodology as used in \cite{wu2017squeezedet}. 
The idle power consumption on P100 and P4000 GPUs are 31 Watts and 5 Watts respectively. We sample the power readings at every 100ms, and once the power consumption becomes stable, we take average power reading as the power consumption of DNN computation. Similarly, we use the average forward pass time (over 200/500 iterations) reported in the Caffe deep learning framework as the inference time of DNN. Further, to mitigate the effect of noise and increase the robustness of power and inference time readings, we repeat the steps mentioned above three times and take an average over these iterations.  Similarly, we use {\tt pcm-power} utility provided by Intel to measure the power on CPU (Section \ref{sec:ApplicabilityOfModel}).

{\bf Energy metrics:} 
``Energy per pixel (EPP)'' (Eq. \ref{eqn:EnergyPerPixel}) measures the energy consumed in processing of one input pixel over entire DNN, whereas ``energy efficiency'' (Eq. \ref{eqn:EnergyEfficinecy}) shows the number of MAC operations performed per unit of energy. We have used EPP  to remove the bias due to differences in input image size used by different DNNs. For example, InceptionV3, InceptionV4, and XceptionNet work with input image size $299\times 299$ while most of DNNs work with input size $224\times 224$ (Table \ref{tab:DataReuse}). 

\begin{align} 
\label{eqn:EnergyPerPixel}
\text{ Energy per pixel (EPP) } = \frac{P_{avg}\times I_t}{\text{\#Pixels in input frame}} \\ 
\label{eqn:EnergyEfficinecy}
\text {Energy Efficiency } = \frac{\text{(batch size)}\times \text{(\# MACs)}}{P_{avg}\times I_t}
\end{align}

\begin{figure*}[htbp]
\centering
\includegraphics[scale=0.45]{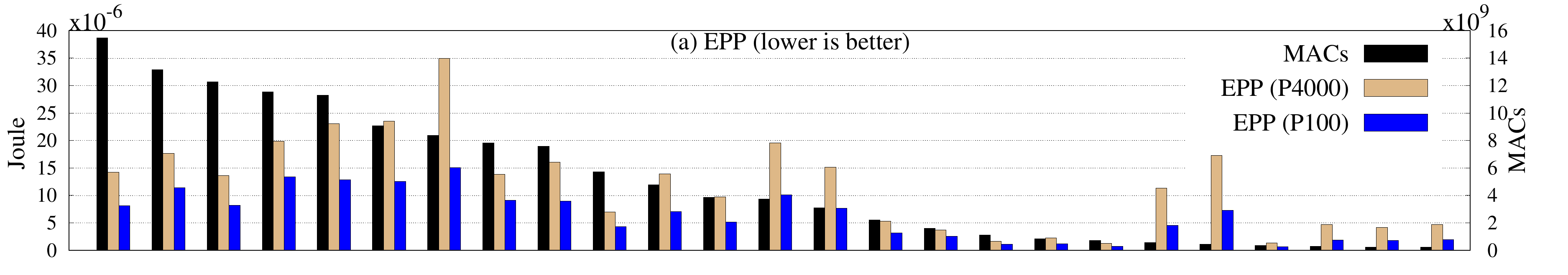}  \\  \vspace{-1.0mm}
\includegraphics[scale=0.45]{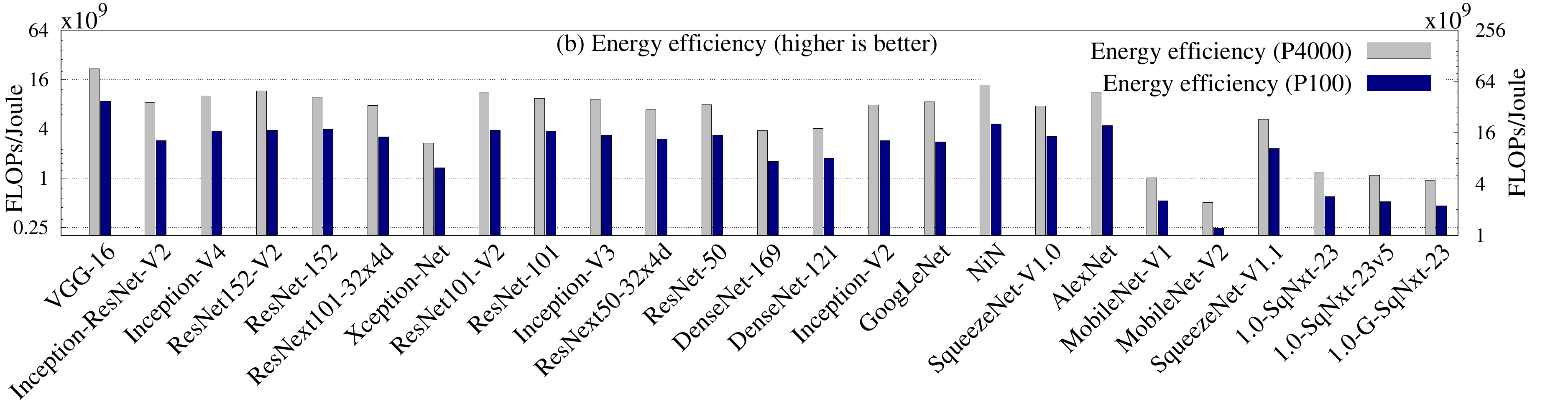}   \\ \vspace{-2.0mm}
\caption{(a) Energy consumption (EPP) and (b) energy efficiency for 25 DNNs, measured on P4000 and P100 GPUs, illustrated in descending order of number of MACs operations in DNNs. (SqNxt=SqueezeNext)}

\label{fig:ExperimentalGraphs}
\end{figure*}

{\bf Correlation coefficients:} We use the Pearson product-moment correlation coefficient (PPMCC) along with Spearman's rank correlation coefficient (SRCC) to examine the relationship between two variables. PPMCC measures the strength of a linear relationship between two variables using the absolute value of data. By contrast, SRCC a nonparametric test that evaluates the monotonicity between two variables using the rank of data, without any presumption about the data distribution \cite{PPMCC,hauke2011comparison}. A value close to +1/-1 for both PPMCC ($r_p$) and SRCC ($r_s$) indicates strong positive/negative correlation whereas a value 0 indicates no correlation between two variables. 

{\em Importance of SRCC:} When we use PPMCC in conjunction with SRCC, we get more insights about the relationship between two variables. For example, since SRCC uses the rank of data instead of absolute values, the effect of outliers is quite meager as compared to that on PPMCC. Therefore, when there is a significant difference between PPMCC and SRCC, either the sample size is insufficient or there exists a group of data points for which a linear/monotonic relationship does not exist. In this paper, we perform experiments on a sufficiently large sample size of 25 DNNs (Section \ref{sec:GeneralityTesting}), and the results with PPMCC and SRCC help find the outliers. Furthermore, the comparative study between PPMCC and SRCC helps analyze the batch size sensitivity of our proposed model.

{\bf Concurrent activations} This includes all the data that needed for the execution of an operation such as convolution \cite{DBLP:journals/corr/abs-1801-04326}. For example, in forward pass, concurrent activation for a convolution operation consist of ifmaps and ofmaps of the current layer, and outputs of previous layers if feed-forward (residual or skip)  connections are present in the network.  The size of maximum concurrent activation depends on types of operation, e.g., different types of convolution (Table \ref{tab:ConvType}), ReLU operation, weight update in back-propagation, etc., and the network topology (linear/non-linear architecture \cite{wang2018superneurons}). Hence, through the inter-layer dependency, which can be obtained from the network's computational graph and the execution order, the size of maximum concurrent data can be estimated.

\section{Energy-efficiency of DNNs} \label{sec:EnergyEfficiency}

Figure \ref{fig:ExperimentalGraphs}(a) shows  EPP and MAC operations, whereas Figure \ref{fig:ExperimentalGraphs}(b) shows energy efficiency of DNNs. EPP and energy efficiency values are measured experimentally on P4000 and P100 GPUs. To make the comparison easier, on X-axis, DNNs are arranged in decreasing order of MAC operations.

{\bf Variation due to depthwise separable convolution:} 
The energy efficiency of MobileNet variants (i.e., MobileNet-V1 and MobileNet-V2) and XceptionNet is significantly lower than that of neighboring DNNs with a higher number of MACs. Hence, their energy per pixel (EPP) is significantly higher than the neighboring DNNs with a higher number of MACs.  In MobileNet variants and XceptionNet, depthwise separable convolution (DWConv, followed by PWConv) has been used to reduce the number of MACs. Both the DWConv and PWConv have a significantly lower degree of data reuse (Table \ref{tab:ConvType}), which results in costlier memory access and makes MACs energy inefficient.

{\bf Variations due to feed-forward/skip connections: } Even though the number of MACs in  MobileNet-V2 is lower than that in MobileNet-V1 and both these networks perform depthwise separable convolution (DWSConv), EPP of the former is significantly higher than that of the latter. This is because MobileNet-V2 has feed-forward connections that increase the concurrent activations and further reduce the already-low data reusability of DWSConv. This translates to low energy efficiency and high EPP of  MobileNet-V2. Similarly, the feed-forward connections in XceptionNet exacerbate the low activation reusability in DWSConv and lower the energy efficiency. Also, XceptionNet has a relatively higher number of MACs, which, in conjunction with low data reusability arising from DWSConv and feed-forward connections, results in the highest EPP among 25 DNNs. In DenseNet models presence of skip connections result in a concatenation of fmaps from previous layers \cite{mlsys2020_15} and increases the number of activations at runtime. Since the number of activations primarily drives the memory-footprint \cite{Jha-VLSID19}, skip connections results in higher memory-footprint, which in turn increases memory access hence reduces energy efficiency.  For the same reason, DenseNet variants also have quite low energy efficiency than their neighboring DNNs, which leads to higher EPP compared to neighboring DNNs with relatively higher MACs. 

{\bf Variation due to low activation reuse:}
Compared to SqueezeNet-V1.1, the variants of SqueezeNext (1.0-G-SqNxt-23, 1.0-SqNxt-23, and 1.0-SqNxt-23v5) have a lower number of MAC operations; however, the EPP of SqueezeNext variants is much higher than that of SqueezeNet-V1.1. The activation reuse in SqueezeNext variants is substantially low compared to that in SqueezeNet-V1.1, and hence, the latter's energy efficiency is substantially higher than the variants of SqueezeNext.

In summary, energy per pixel of a DNN depends on the total number of MAC operations and the energy consumed by each MAC operation. Former can easily be estimated at the design time, but estimating the latter is difficult due to varying degrees of data reuse in different layers of DNNs.

\section{Conventional Approaches } \label{sec:PreMethodology}

We discuss two possible approaches for estimating the data reuse in DNNs and also show their limitations. These approaches are (1)  layer-wise arithmetic intensity (Section \ref{subsec:LayerWiseAIMetric}) and (2)  cumulative arithmetic intensity ($AI_c$) (Section \ref{subsec:CumulativeAI}). 
 
\subsection{Layer-wise arithmetic intensity}\label{subsec:LayerWiseAIMetric}
\begin{figure}[htbp]
\centering
\includegraphics[scale=0.32]{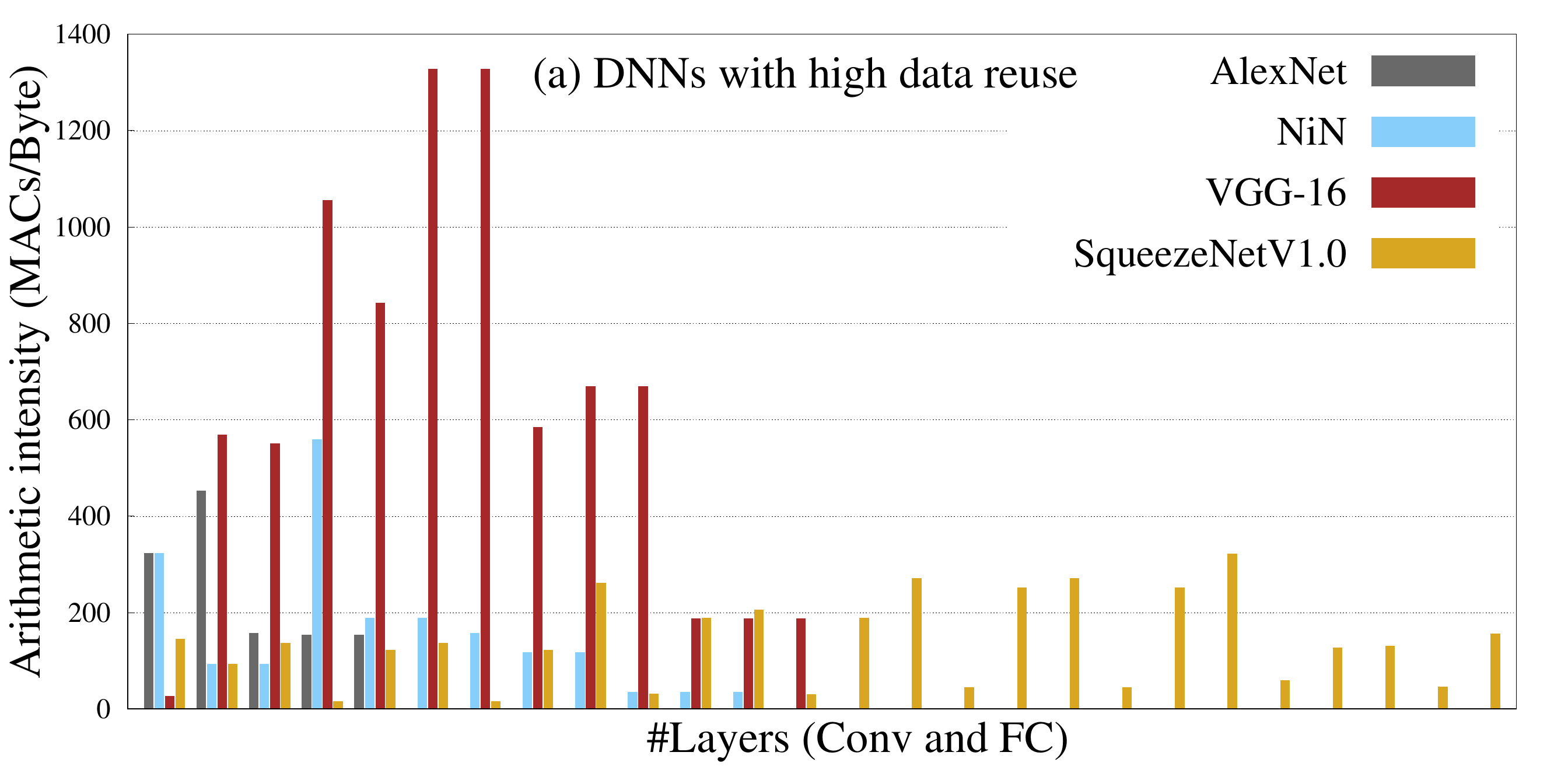}  
\includegraphics[scale=0.32]{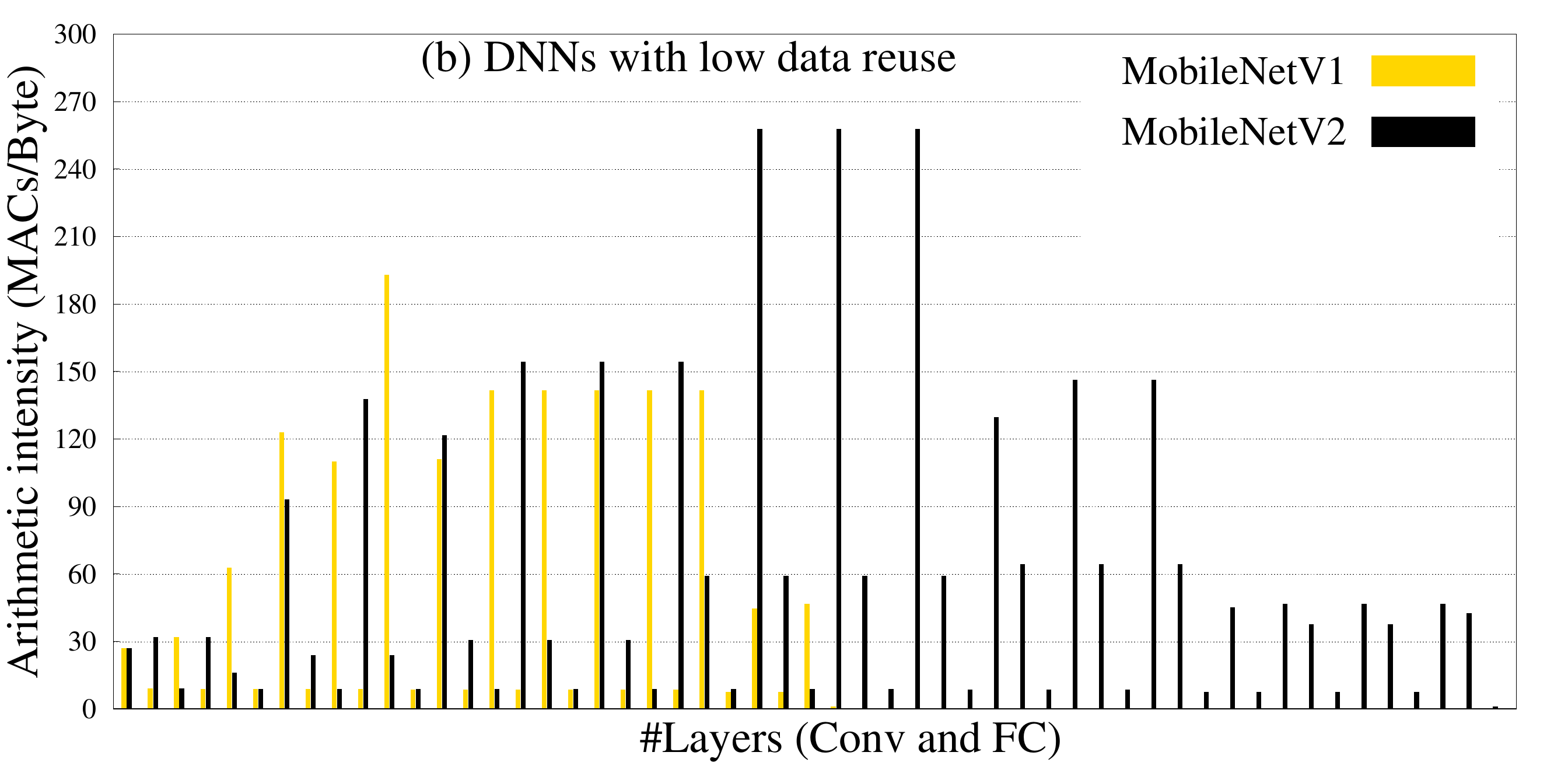} 

\caption{ Layer-wise arithmetic intensity in DNNs with high data reuse (top) and  DNNs with low data reuse (bottom).}
\label{fig:LayerWiseAI}
\end{figure}

Figure \ref{fig:LayerWiseAI} shows the arithmetic intensity of a layer (Conv and FC) defined as the ratio of ``number of MACs performed in that layer'' to  ``the sum of the total number of weights and activations in that layer''. We divide DNNs in two categories based on their degree of data reuse: those with higher data reuse, e.g., AlexNet, VGG-16, NiN, and SqueezeNetV1.0 (Figure \ref{fig:LayerWiseAI}(a)) and those with lower data reuse, e.g., MobileNet-V1 and MobileNet-V2 (Figure \ref{fig:LayerWiseAI}(b)).
From Figure \ref{fig:LayerWiseAI}(a), we observe that arithmetic intensity of nearly all layers of VGG-16 and NiN are higher and lower (respectively) than that of other DNNs. Similarly, from Figure \ref{fig:LayerWiseAI}(b), the layer-wise arithmetic intensity in MobileNet-V2 is comparable or higher than that of the MobileNet-V1.

Table \ref{tab:LayerWiseAI} shows the median and variance of the arithmetic intensity of layers in these DNNs. The median value of layer-wise arithmetic intensity in VGG-16 is significantly higher than the median value of layer-wise arithmetic intensity in other DNNs with higher data reuse. Also, NiN has the lowest median among the DNNs with higher data reuse.  Intuitively, the energy efficiency of VGG-16 should be higher compared to other DNNs with high data reuse, and also, the energy efficiency of NiN should be lowest in the same group. Surprisingly, on both P4000 and P100 GPUs, \textit{the energy efficiency of NiN is highest, and that of the VGG-16 is lowest among the DNNs with higher data reuse}. 

\begin{table} [htbp] 
\caption{Median and variance of layer-wise arithmetic intensities in DNNs (SqNet=SqueezeNetV1.0)}
\label{tab:LayerWiseAI}\centering 
\resizebox{0.48\textwidth}{!}{
\begin{tabular}{ |c|c|c|c|c|c|c| } \hline
 \textbf{Metric} & \textbf{AlexNet} & \textbf{VGG-16} & \textbf{NiN}  & \textbf{SqNet} & \textbf{MobileNet-V1} & \textbf{MobileNet-V2} \\ \hline 
 
 Median  & 154    &560 &    117    &134&    18&    32   \\ \hline
 Variance   &     2.69E+04&    2.09E+05&    2.24E+04&    8.30E+03&    3.70E+03&    4.52E+03 \\ \hline

\end{tabular} }
\end{table}

Similarly, MobileNet-V2 has lower energy efficiency than MobileNet-V1, which is counter-intuitive. By observing the variance of the layer-wise arithmetic intensity of DNNs in Table \ref{tab:LayerWiseAI}, one may argue that since the variance values of VGG-16 and MobileNet-V2 are higher than the other DNNs in their respective groups, hence their energy efficiency is lowest, despite having higher median. This argument is flawed because it cannot explain why NiN has higher energy efficiency than SqueezeNetV1.0, even when NiN has a significantly higher variance than SqueezeNetV1.0. The above discussion proves that \textit{``layer-wise arithmetic intensity'' is not a good indicator of energy efficiency of DNNs}  with both high and low data reuse. Besides this, finding layer-wise arithmetic intensity in DNNs such as Inception-V3, Inception-V4, and Inception-ResNet-v2, is tedious because these networks have thousands of Conv layers and many layers have several branches leading to irregular computation and reuse patterns.

\subsection{Cumulative arithmetic intensity } \label{subsec:CumulativeAI}
\begin{figure*} [htbp] \centering
\begin{subfigure}{0.49\textwidth}
\includegraphics[scale=0.48]{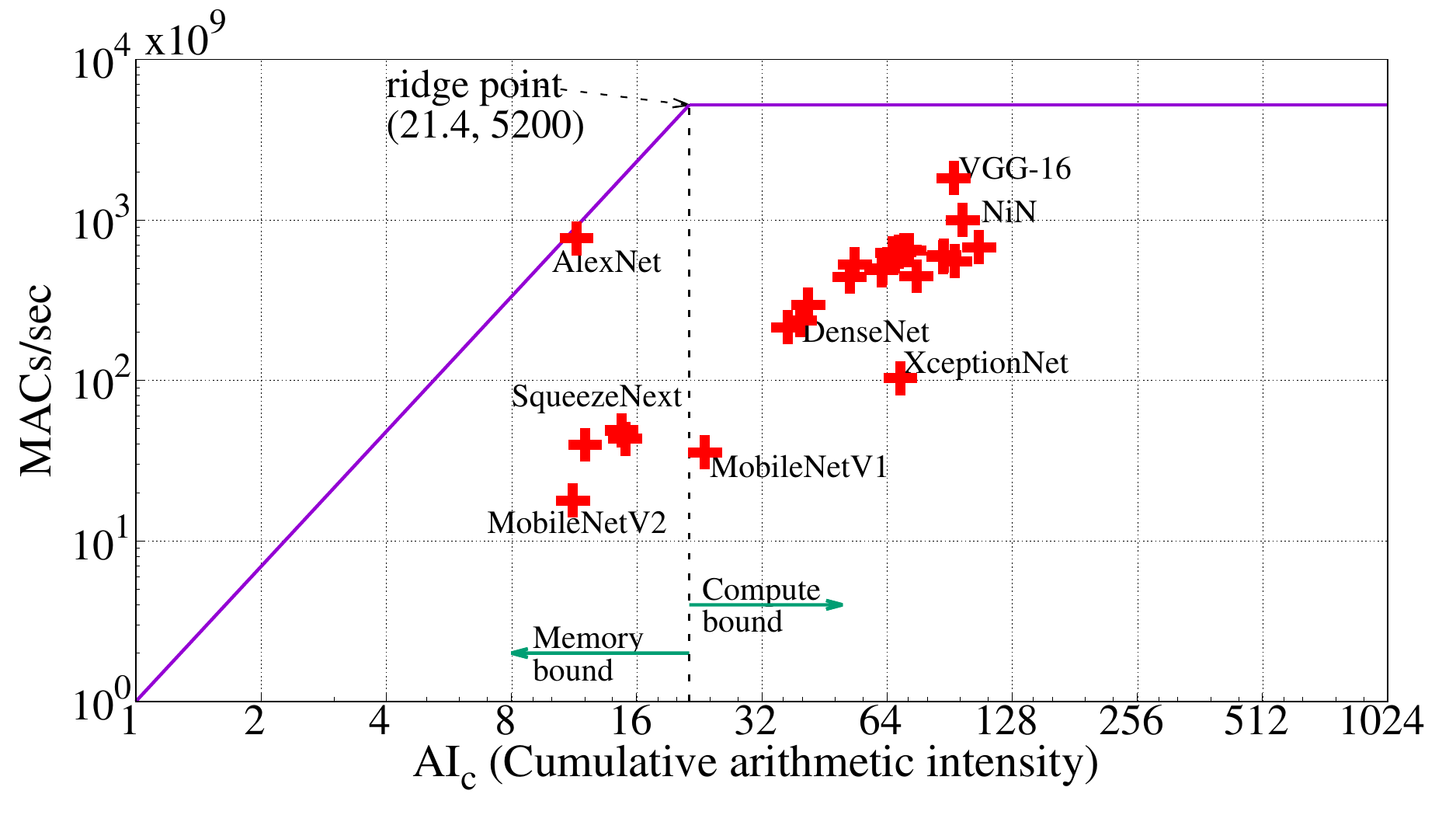}
\caption{Roofline model (with $AI_c$) on P4000 GPU} \label{fig:RooflineP4000}
\end{subfigure}
%\hspace*{\fill} % separation between the subfigures
\begin{subfigure}{0.49\textwidth}
\includegraphics[scale=0.48]{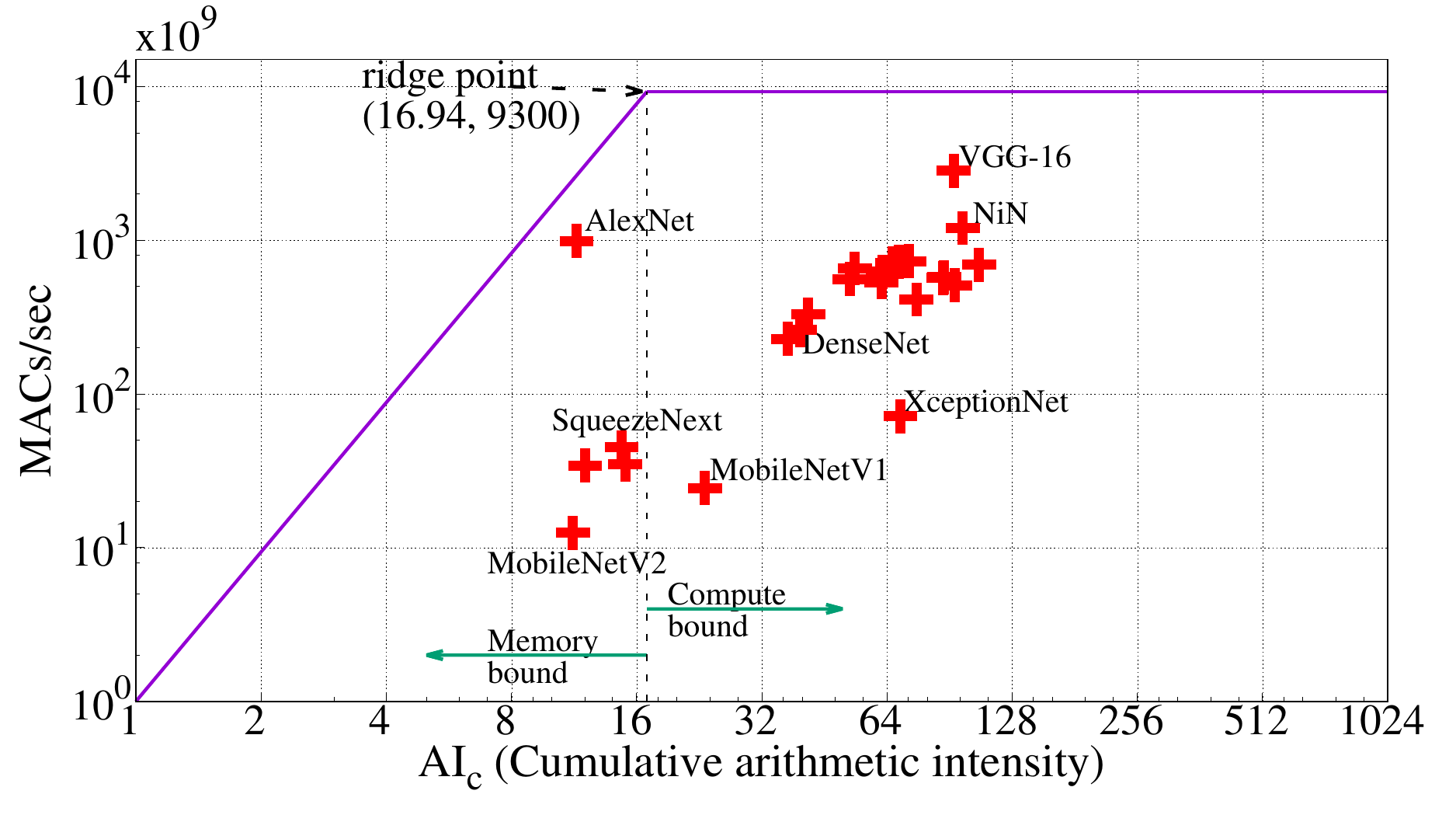}
\caption{Roofline model (with $AI_c$) on P100 GPU} \label{fig:RooflineP100}
\end{subfigure} \\
%\hspace*{\fill} % separation between the subfigures
\begin{subfigure}{0.49\textwidth}
\includegraphics[scale=0.48]{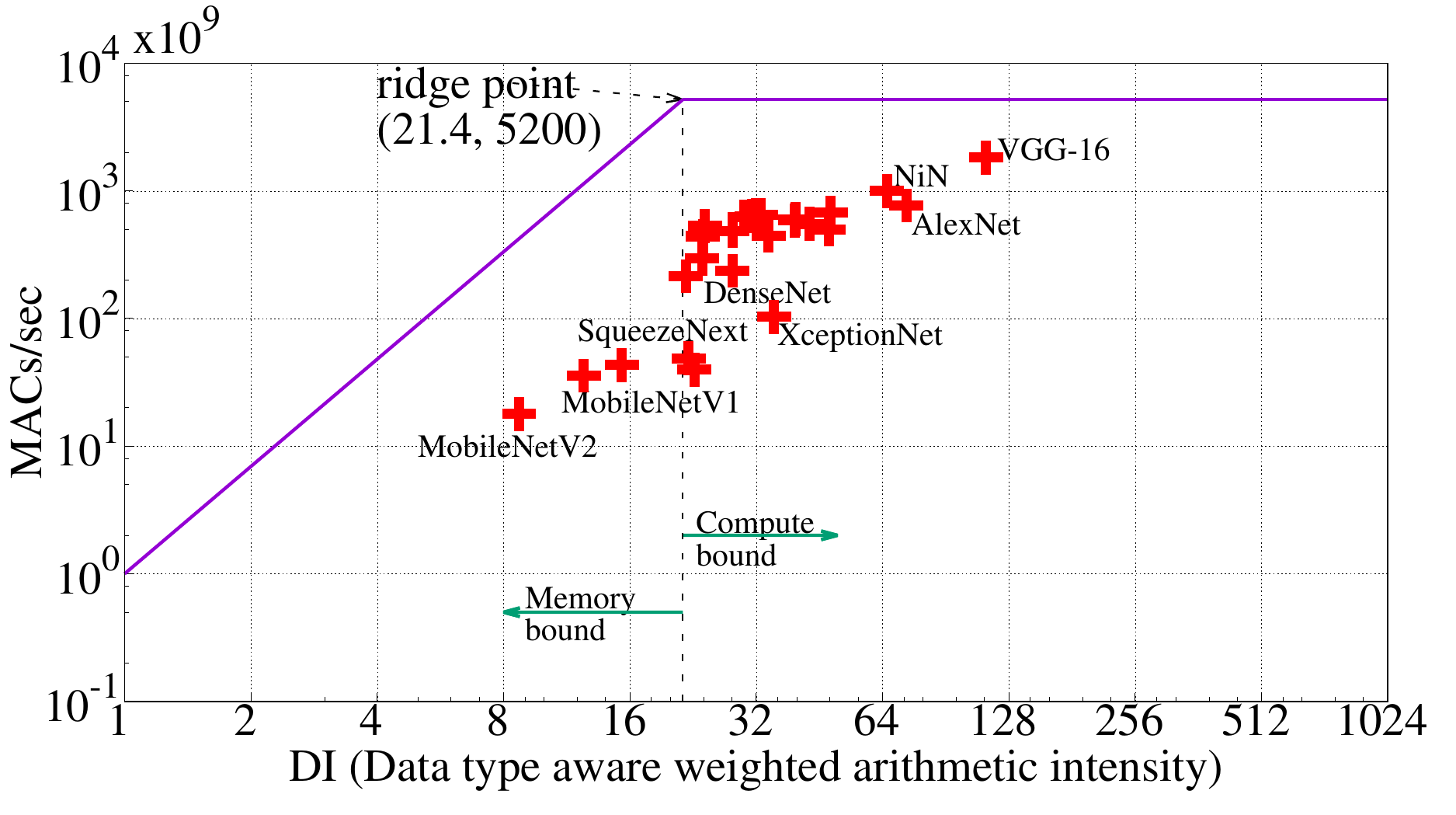}
\caption{Roofline model (with $DI$) on P4000 GPU} \label{fig:RooflineAIwP4000}
\end{subfigure}
\begin{subfigure}{0.49\textwidth}
\includegraphics[scale=0.48]{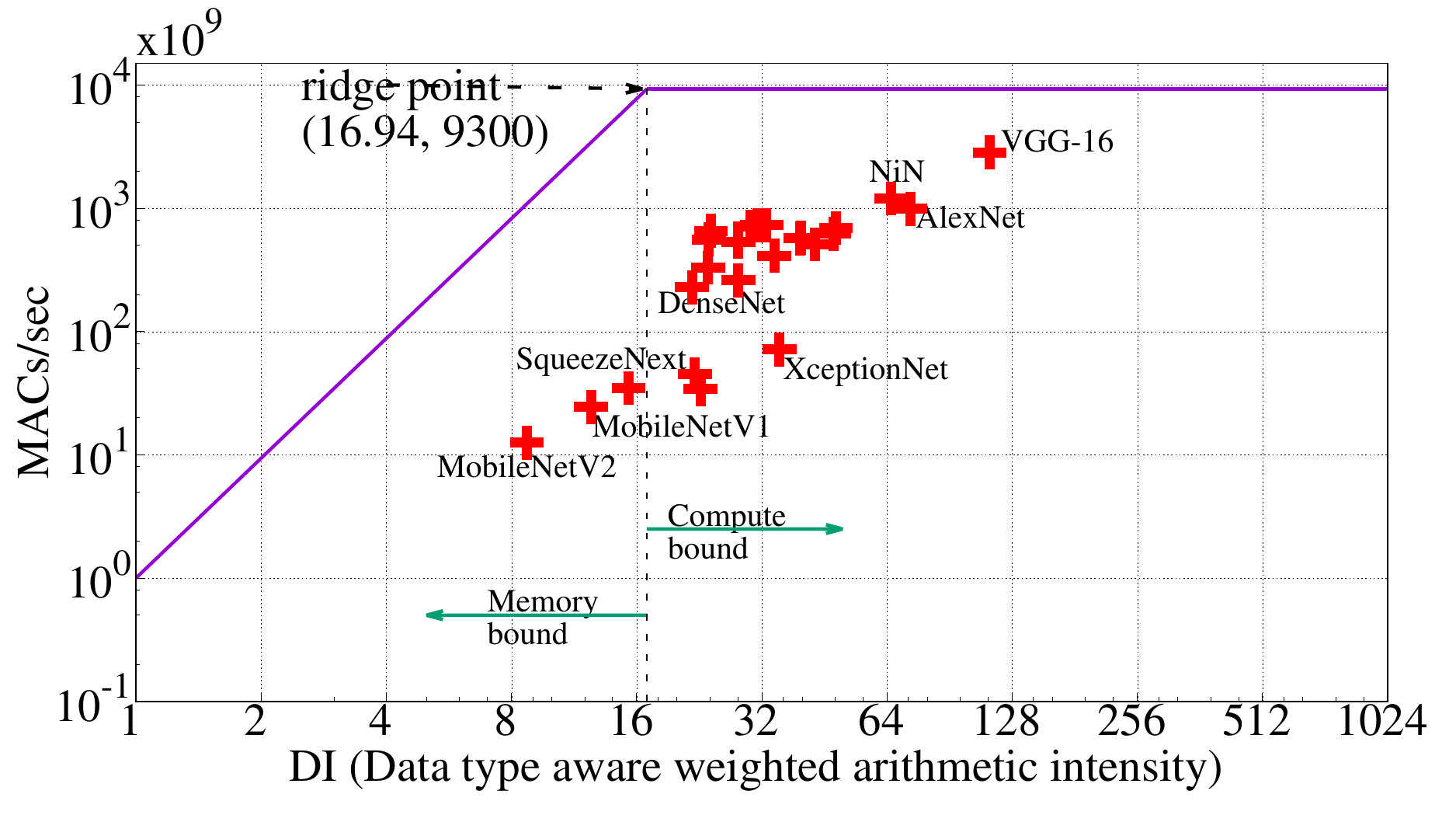}
\caption{Roofline model (with $DI$) on P100 GPU} \label{fig:RooflineAIwP100}
\end{subfigure} \\
\begin{subfigure}{0.49\textwidth}
\includegraphics[scale=0.48]{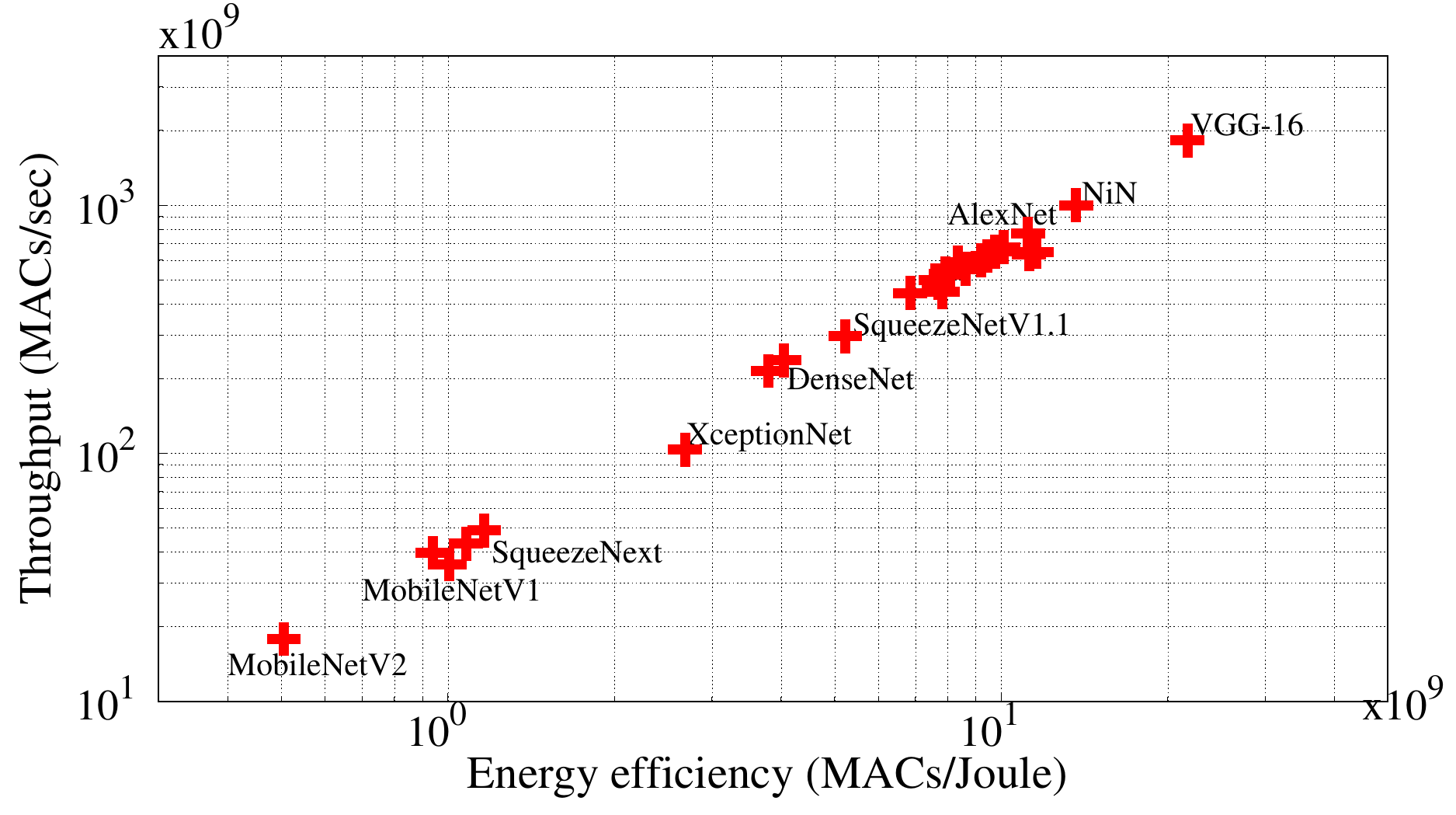}
\caption{Throughput and energy-efficiency on P4000 GPU} \label{fig:RooflineCorrelationP4000}
\end{subfigure}
\begin{subfigure}{0.49\textwidth}
\includegraphics[scale=0.48]{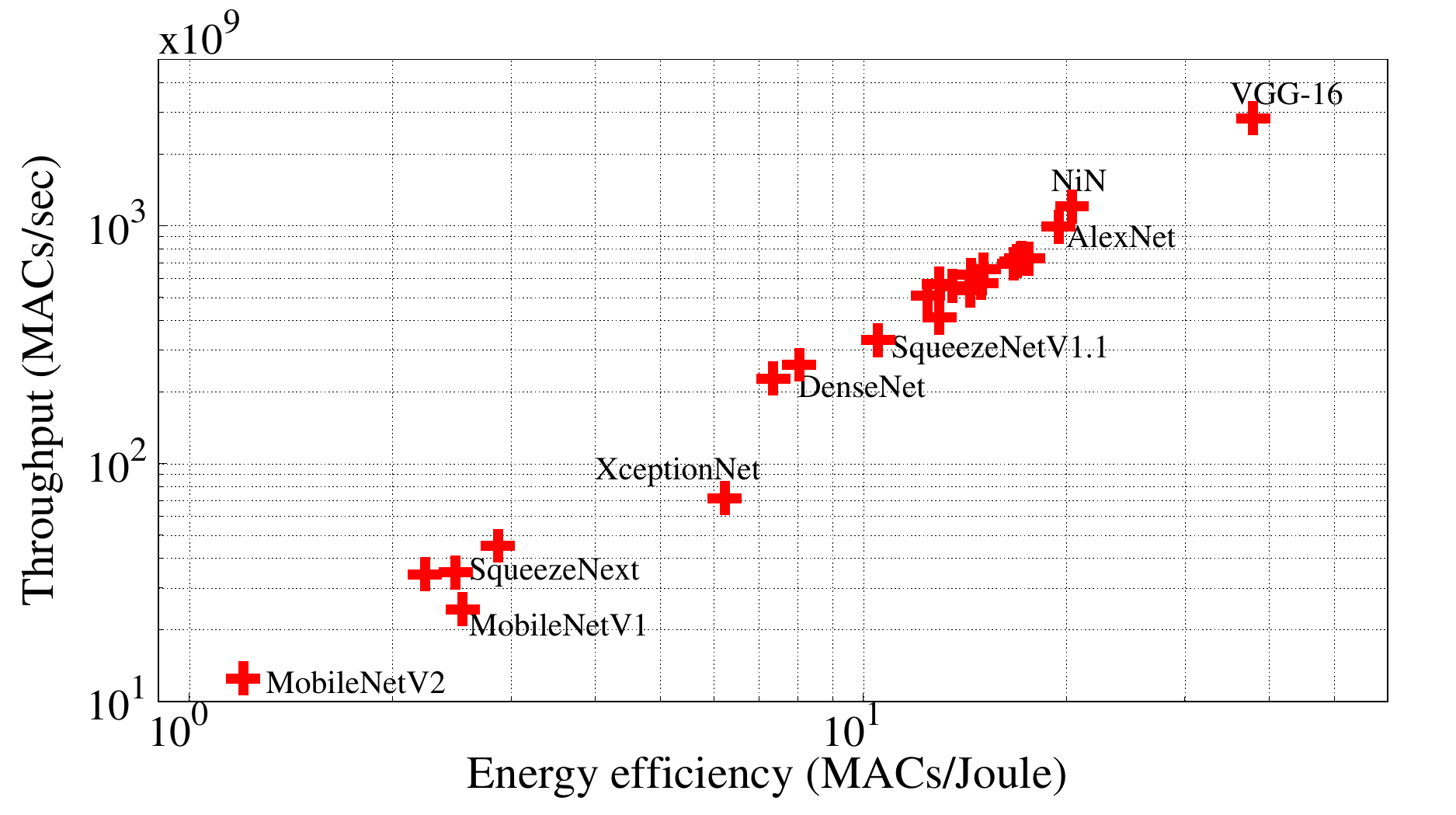}
\caption{Throughput and energy-efficiency on P100 GPU} \label{fig:RooflineCorrelationP100}
\end{subfigure}
\caption{Roofline model with (a) conventional metric ($AI_c$), and (c) our proposed metric ($DI$) on x-axis; and (e) energy efficiency measured on P4000 GPU for 25 DNNs. (b), (d), (f): these results on P100 GPU. All measured values are with $B$=4}
\label{fig:Roofline}
\end{figure*}

For each DNN, a single metric, viz.,  arithmetic intensity ($AI_c = \frac{M_c}{W + A} $) is  defined as the ratio of ``total number of MAC operations performed by network in one forward pass'' to ``the sum of total number of weights and activations that network has''. We have plotted the roofline models with $AI_c$ for 25 DNNs (Table \ref{tab:DataReuse}) on two GPUs P4000 (Figure \ref{fig:Roofline}(a)) and P100 (Figure \ref{fig:Roofline}(b)).   We also measure the energy efficiency of 25 DNNs on P4000 and P100 and plot in Figures \ref{fig:Roofline}(e) and  \ref{fig:Roofline}(f), respectively.

In the roofline models (with $AI_c$) on both GPUs (Figure \ref{fig:Roofline}(a) and (b)), MobileNet-V1, DenseNet and XceptionNet are predicted as compute-bound whereas AlexNet is predicted as bandwidth-bound. It is well known that due to the costlier off-chip accesses, bandwidth-bound operations are energy inefficient compared to compute-bound operations. Despite this, AlexNet has substantially higher energy efficiency than MobileNet-V1, DenseNet, and XceptionNet on both the   GPUs (refer Figure \ref{fig:Roofline}(e) and \ref{fig:Roofline}(f)). 
Also, XceptionNet and DenseNet are predicted as compute-bound, but their energy efficiency is lower than that of the AlexNet. In summary, {\em data reuse predicted by $AI_c$ is not correlated with the energy efficiency of DNNs}.

To understand the reason behind limitations of $AI_c$, we study the architecture of XceptionNet and  DenseNet. We found that DenseNet has many skip connections which results in concatenation of fmaps from previous layers. This concatenation of fmaps increases the number of concurrent activations \cite{mlsys2020_15,DBLP:journals/corr/abs-1801-04326} and decreases the effective data reuse. Similarly, XceptionNet uses DWSconv which has very low data reuse (Table \ref{tab:RelativeValueOfConvType}). It also has skip connections which increase the concurrent activation data and further reduce the data reuse. $AI_c$ \big($=\frac{M_c/W}{1 + A/W} = \frac{M_c/A}{1 + W/A}$\big) gives equal importance to weight reuse ($M_c$/$W$) and activation reuse ($M_c$/$A$) and hence, it is unable to capture the runtime change in data reuse.

\begin{figure*}[htbp]
\centering
\includegraphics[scale=0.45]{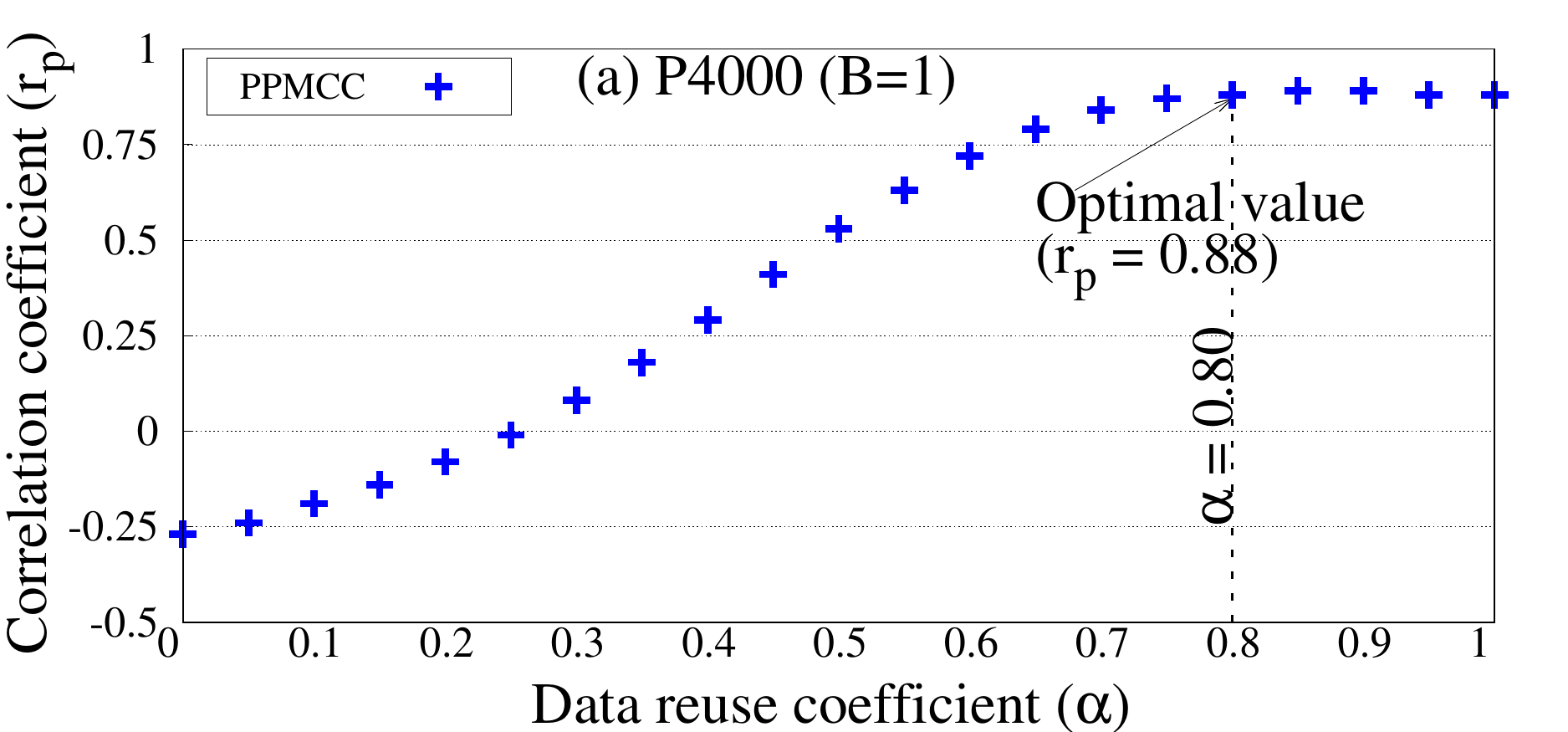}   
\includegraphics[scale=0.45]{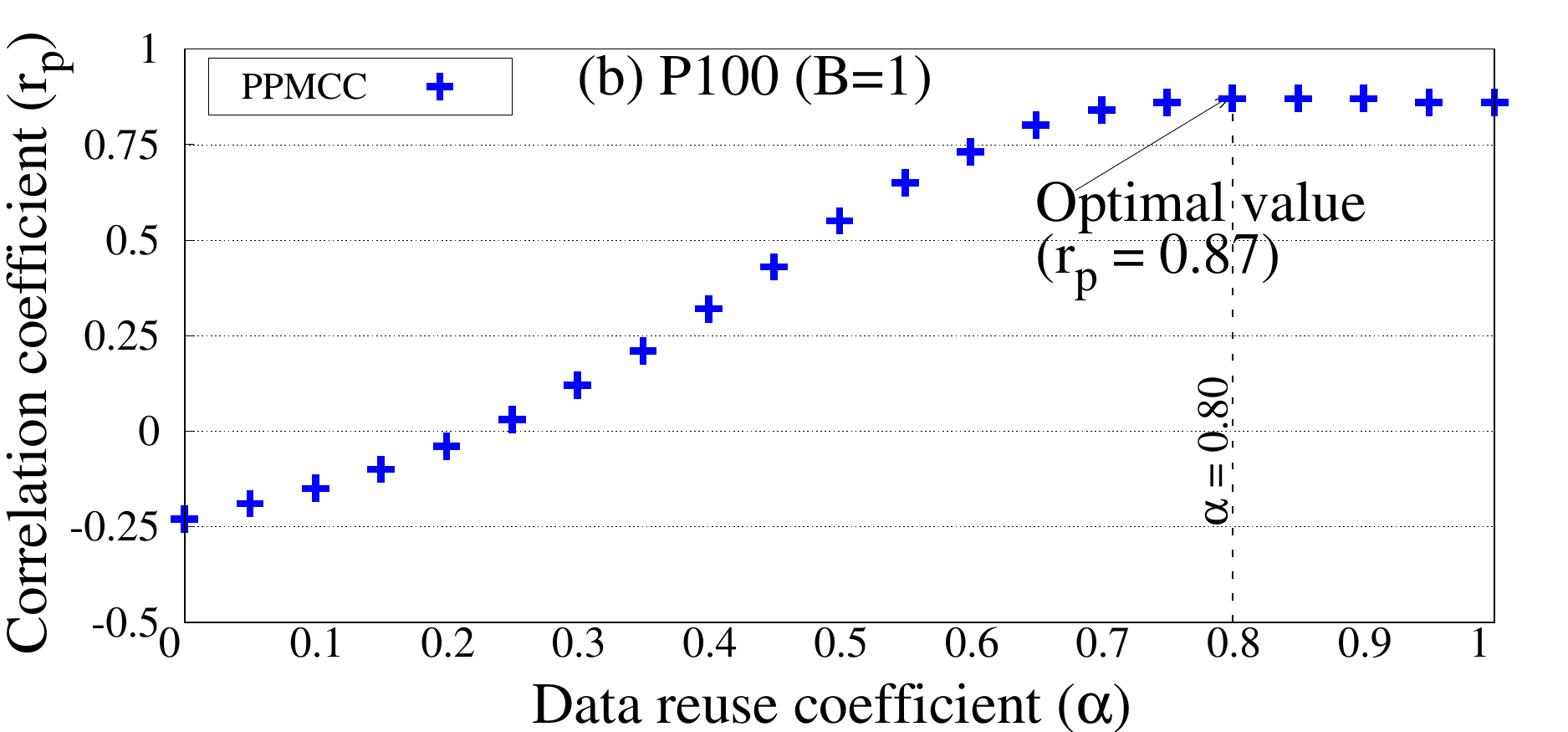} 
\includegraphics[scale=0.45]{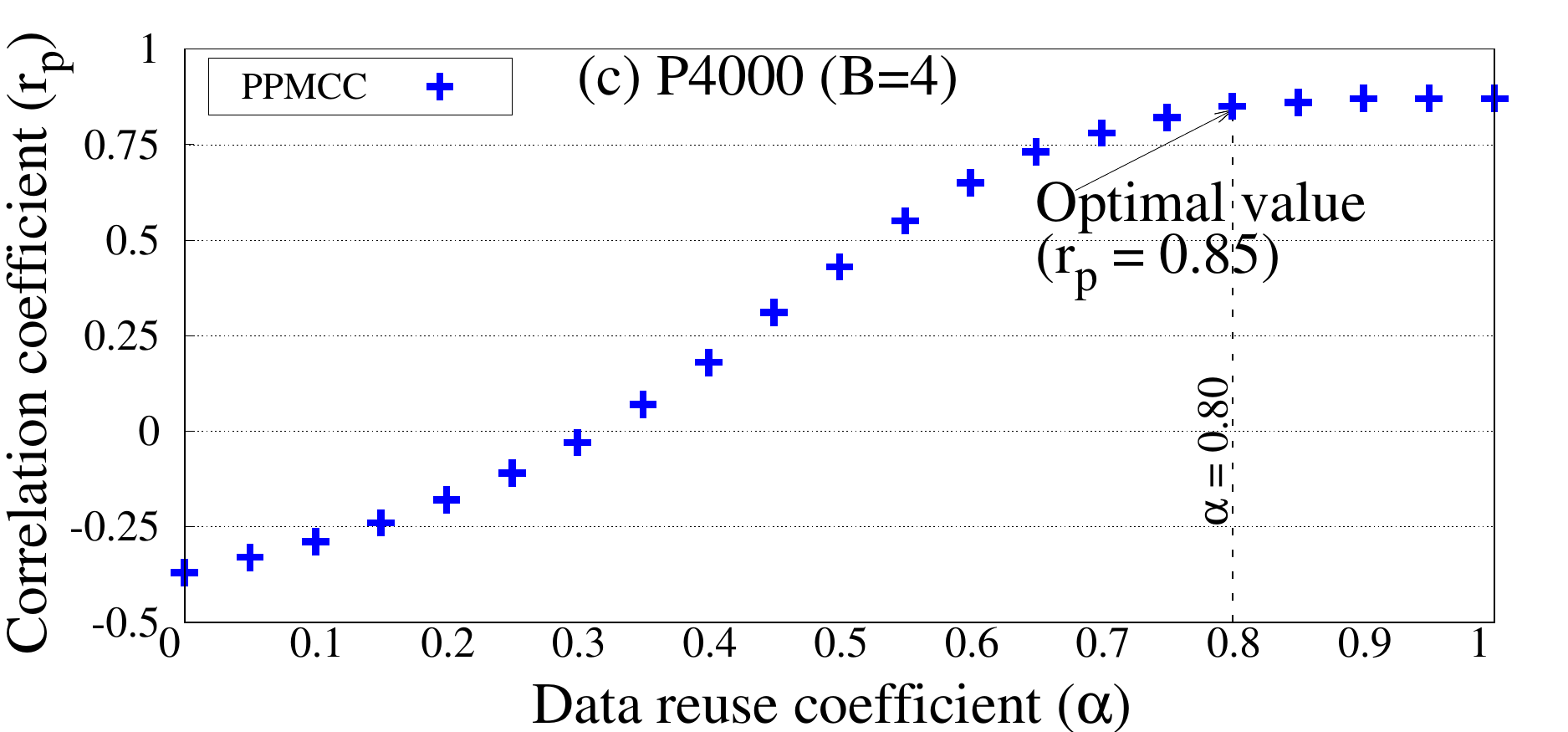}   
\includegraphics[scale=0.45]{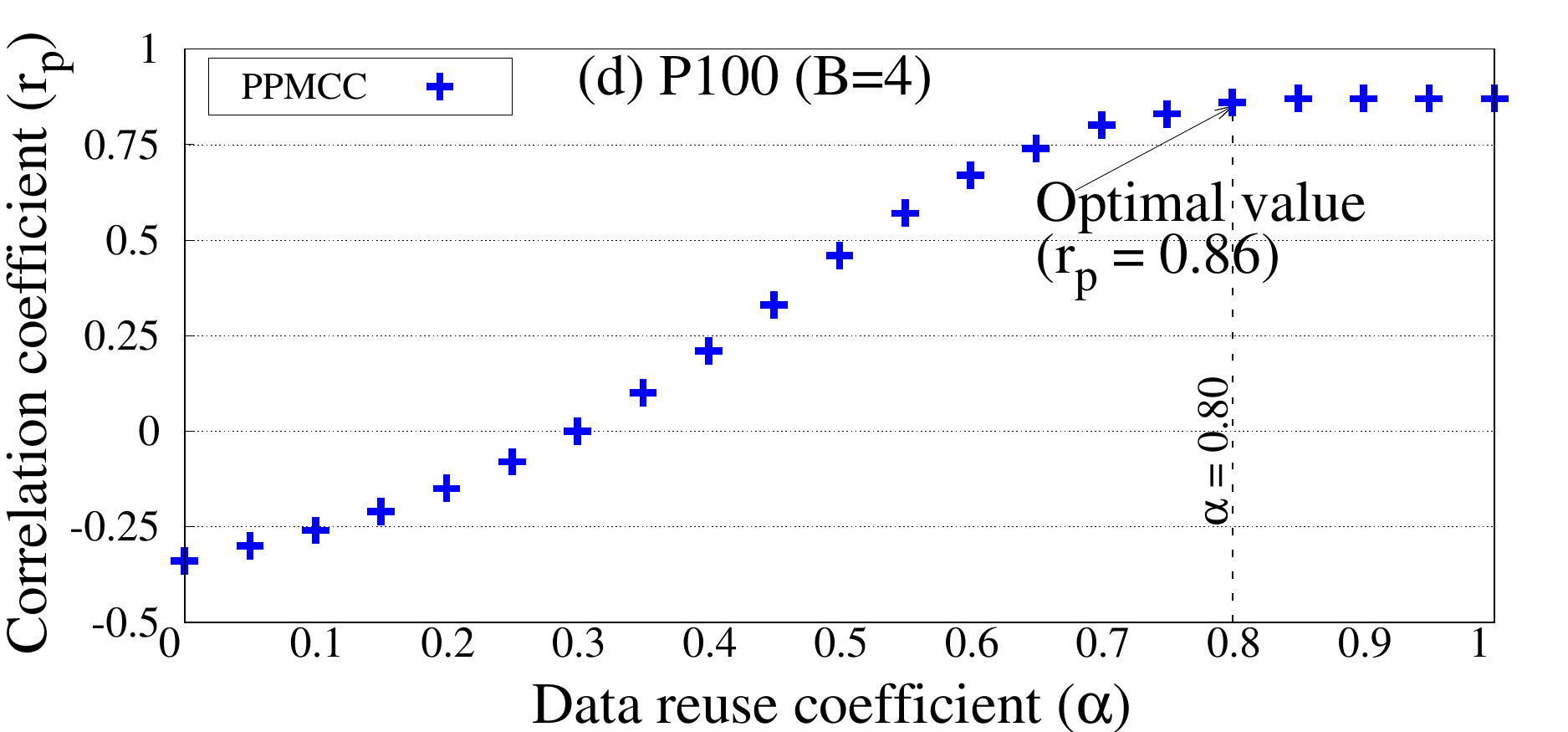} 
\caption{Variations in the PPMCC calculated between weighted arithmetic intensity at different values of $\alpha$, and energy efficiency measured for 25 DNNs on P4000 GPU with $B$=1 (a), and $B=4$ (c). (b) and (d): these results on P100 GPU}
\label{fig:PearsonGPU}
\end{figure*}

\section{Proposed Model} \label{sec:methodology}

In this section, we first discuss the need to give unequal importance to the reuse of different data types, specifically weights and activations (Section \ref{subsec:WhyUnequal}). We then propose a model that more accurately incorporates the dynamics of data reuse in DNNs (Section \ref{subsec:MethodologyAIcomparison}). We highlight the effectiveness of our model (Section \ref{sec:SalientFeatures}) and compare it with $AI_c$ to get more insights and explain why $AI_c$ fails to predict the nature (memory-bound/compute-bound) of some DNNs. 

\subsection{Reuse of different data types has unequal importance} \label{subsec:WhyUnequal}

As we know, the number of weights ($W$) in a DNN does not change at runtime. However, the number of concurrent activations can change at run time and grows in proportion to the number of feed-forward connections (as explained in Section \ref{sec:setup}). DNNs such as DenseNet have a relatively higher number of skip connections, which leads to a substantial increase in concurrent activations. Further, as shown in Table \ref{tab:DataReuse}, the ratio of the total number of activations to the total number of weights, i.e., $\frac{A}{W}$ varies significantly across different DNNs, ranging from 0.03 in AlexNet to 32.80 in 1.0-G-SqNxt. This imbalance between $W$ and $A$ creates an imbalance between $\frac{M_c}{W}$ and $\frac{M_c}{A}$, for example, compact DNNs such as MobileNet and SqueezeNext have very low $\frac{M_c}{A}$ (Table \ref{tab:DataReuse}). To account for the imbalance between the weight and activation reuse and also to model the runtime change in effective data reuse, our model decouples the weight and activation reuse.

\textbf{Does a metric that decouples weight and activation reuse inherit the properties of arithmetic intensity?} As shown in Table \ref{tab:ConvType}, the arithmetic intensity of commonly used convolution types are different. On decoupling the data reuse in terms of weight and activation reuse, we find that in all convolution types, the weight reuse is the same (i.e., $S_{o}^2$), whereas activation reuse is different. {\em This is quite interesting because it shows that the variation in activation reuse governs the variation in arithmetic intensity for different convolution types}. Since lower arithmetic intensity leads to higher bandwidth pressure, lower activation reuse leads to a higher number of memory accesses, making DNN energy inefficient. Based on these insights,  activation reuse should be given more importance than weight reuse for computing the overall data reuse. As shown in  Table \ref{tab:RelativeValueOfConvType}, the activation reuse decreases from standard convolution to DWConv in the same order in which the relative value of arithmetic intensity is decreasing (but with different magnitudes). This confirms that {\em a metric that decouples weight and activation reuse inherits the properties of arithmetic intensity. }

\subsection{Decoupling the weight and activation reuse} \label{subsec:MethodologyAIcomparison}

We now decouple weight, and activation reuse from the formulation of $AI_c$ and establish a relation between arithmetic intensity ($AI_c$) and weight and activation reuse. 
Since arithmetic mean is never less than harmonic mean, we have

\begin{equation}
\frac{W + A}{2} \geqslant \frac{2\times W\times A}{W + A} \nonumber  \implies 
\frac{2\times M_c}{W + A} \leqslant \frac{M_c \times (W + A)}{2\times W\times A} \nonumber 
\end{equation}
\begin{equation}
\implies
\frac{M_c}{W + A} \leqslant \frac{1}{4}\times \bigg[ \frac{M_c}{A} + \frac{M_c}{W} \bigg] \nonumber 
\end{equation}
\begin{equation}
\implies
\label{eqn:TheoreticalAIComp}
AI_c \leqslant \frac{1}{4}\times [ \text{ActivationReuse} + \text{WeightReuse}]
\end{equation}

To give unequal importance, we introduce a {\em data reuse coefficient} ($\alpha$) in Eq. \ref{eqn:TheoreticalAIComp}, where $0 \leq \alpha \leq 1$. Different values of $\alpha$ would give different weightage  to both types of reuse. 
After introducing $\alpha$ in Eq. \ref{eqn:TheoreticalAIComp}, we refer  resultant metric as ``data type aware weighted arithmetic intensity'' ($DI$).

\begin{align}
\label{eqn:WeightedAI}
DI =  \frac{[\alpha\times \text{ActivationReuse} + (1 - \alpha)\times \text{WeightReuse}]}{4}
\end{align}

\begin{figure*}[htbp]
\centering
\includegraphics[scale=0.45]{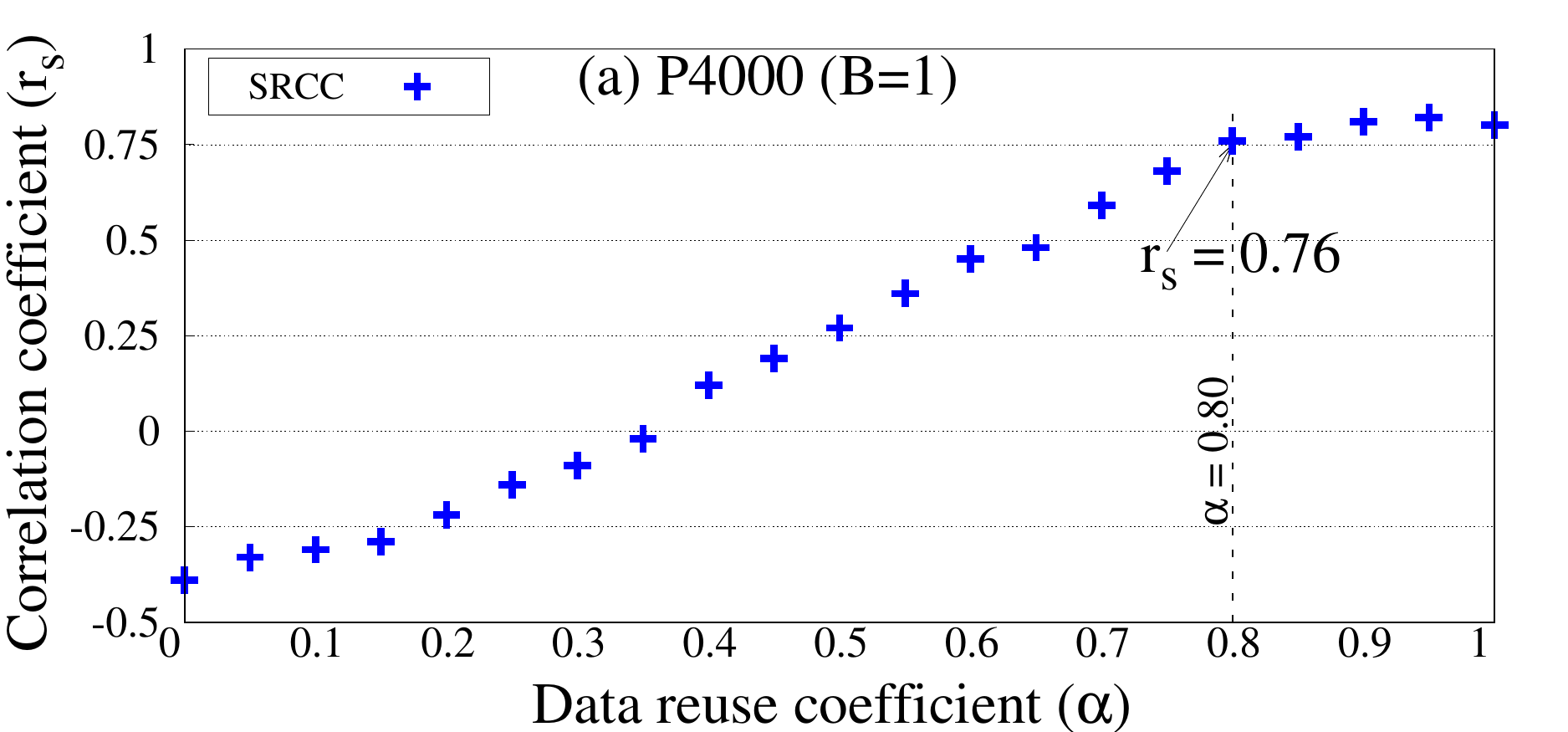}   
\includegraphics[scale=0.45]{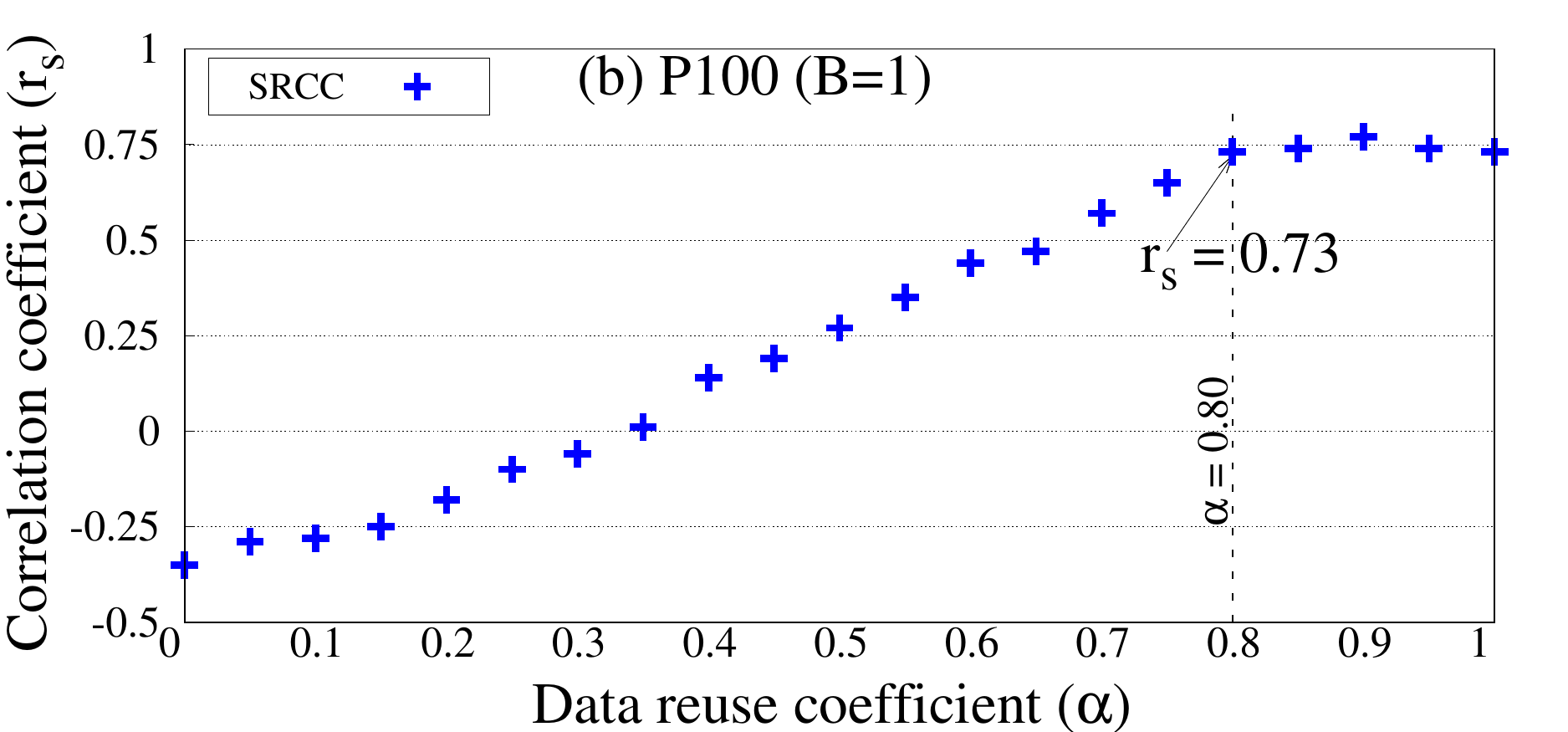} 
\includegraphics[scale=0.45]{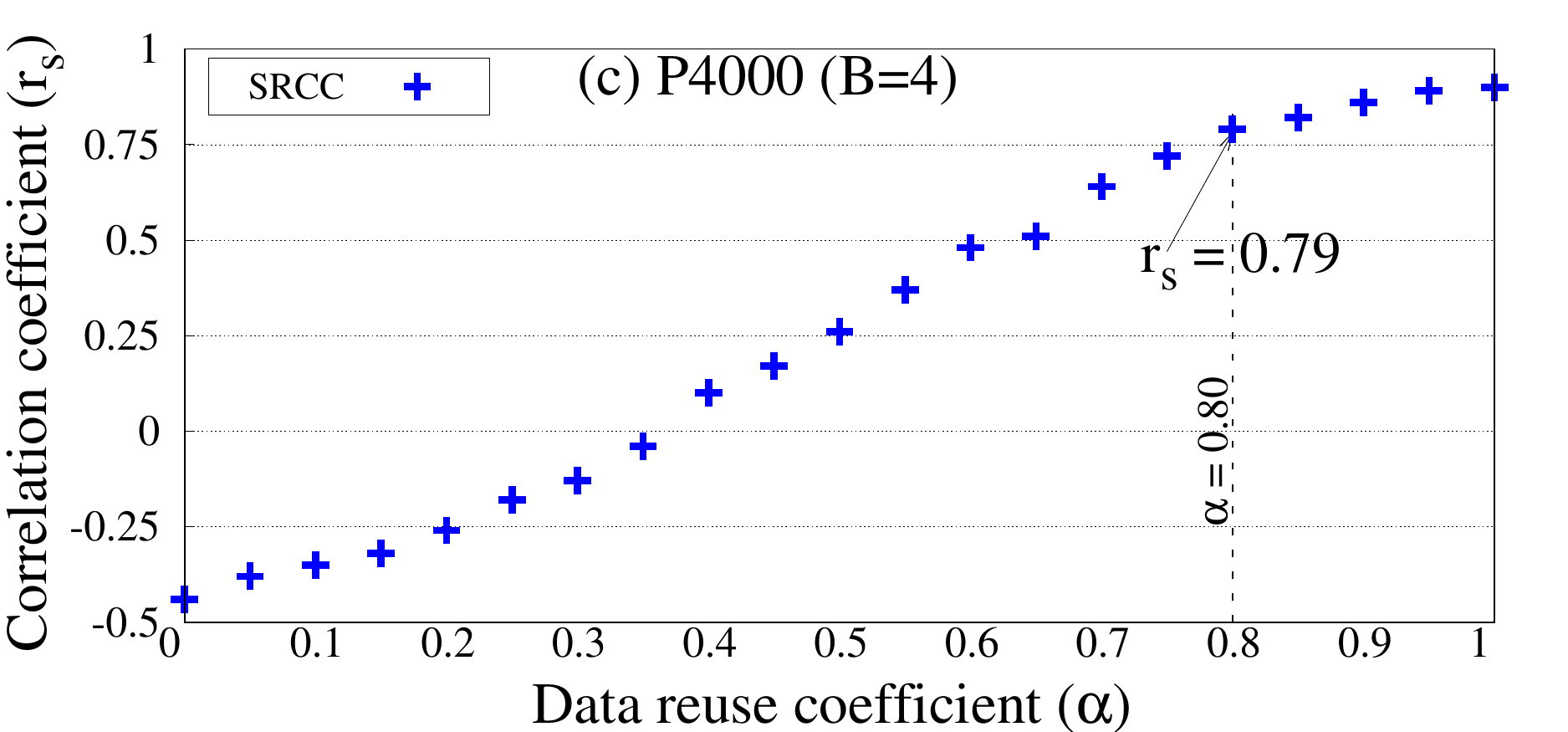}   
\includegraphics[scale=0.45]{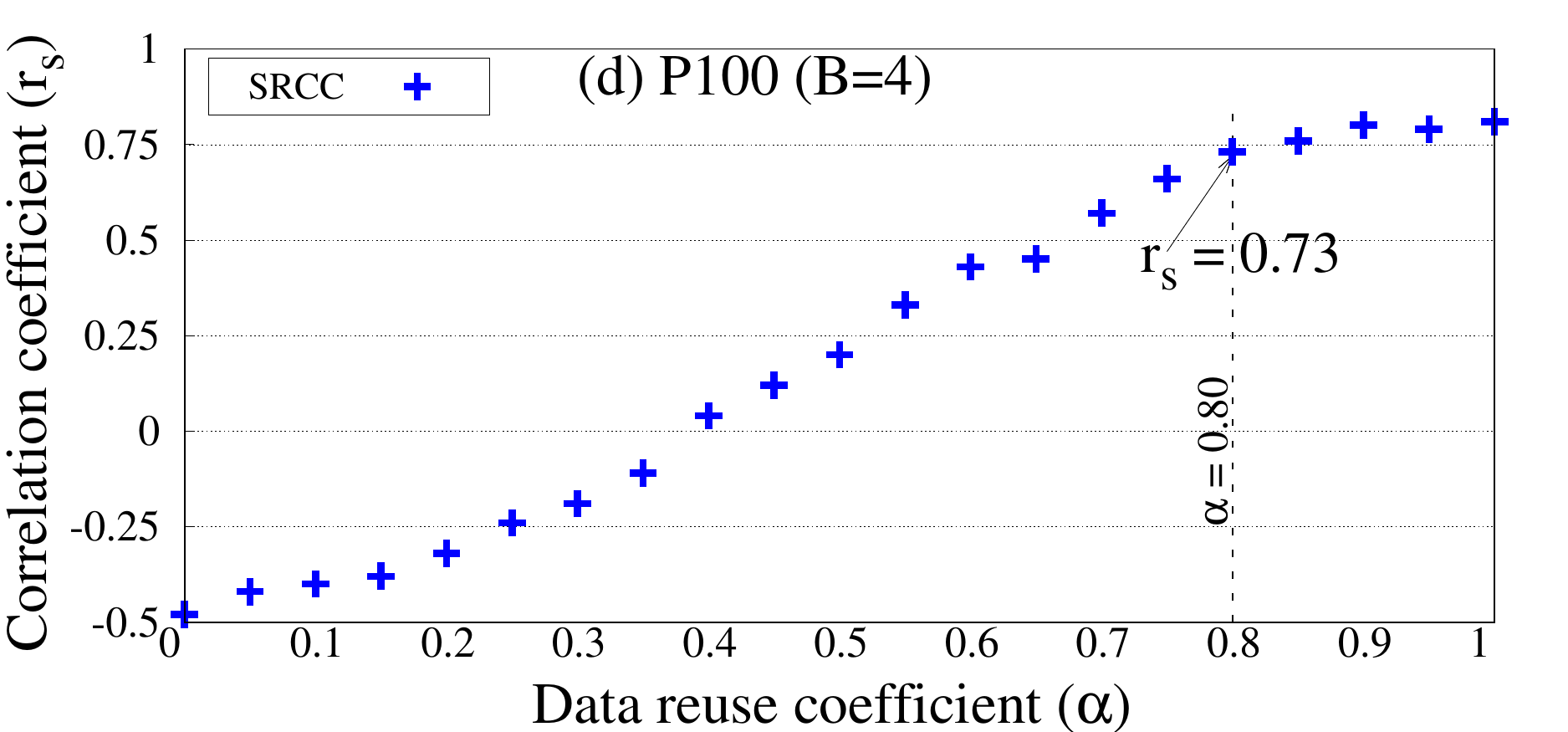} 
\caption{Variations in the SRCC calculated between weighted arithmetic intensity at different values of $\alpha$, and energy efficiency measured for 25 DNNs on  P4000 GPU with $B$=1 (a), and $B=4$ (c). (b) and (d): these results on P100 GPU}
\label{fig:SpearmanGPU}
\end{figure*}

To find the value of $\alpha$ such that $DI$ would become a more accurate indicator of energy efficiency, we find the Pearson correlation coefficient ($r_p$) between $DI$ and energy efficiency for $\alpha$ values between 0 to 1 with a step size of 0.05 (Fig. \ref{fig:PearsonGPU}). We experimentally measure the energy efficiency of 25 state-of-the-art DNNs on both the GPUs P4000 and P100. Figure \ref{fig:PearsonGPU} and Figure \ref{fig:SpearmanGPU} show the results. We observe that on both GPUs, with increasing value of $\alpha$, the correlation ($r_p$) continues to increase till it reaches a plateau at $\alpha \approx 0.80$ (Figure \ref{fig:PearsonGPU}). However, the correlation ($r_s$) does not saturate at $\alpha$=0.8 and keeps increasing even with the higher values of $\alpha$ (Figure \ref{fig:SpearmanGPU}). This difference between the saturation points indicates that there exist few DNNs for which the linear relationship between the weighted arithmetic intensity and energy-efficiency does not hold at higher  $\alpha$. Hence, we take $\alpha =0.80$ and substitute this value in Eq. \ref{eqn:WeightedAI} that gives the final expression for $DI$. Note that the trends in the increase in both the $r_p$ and $r_s$ with the higher values of $\alpha$ are consistent on both GPUs, which have significantly different compute power and CMR (Table \ref{tab:GPUsFeatures}). Thus, {\em the proposed metric $DI$ is platform-agnostic}. Also, at $\alpha$=0.8 the absolute value of $r_p$ is higher than that of the $r_s$. Hence, even if there are outliers in the data (which affect $r_p$ more than $r_s$), the linear association between $DI$ and energy-efficiency exists for those outliers too.

\subsection{Batch Size Sensitivity of Our Proposed Model} \label{subsec:BatchSizeSensitivity}

To study the effect of batch size on our proposed model ($DI$ with $\alpha$ = 0.80) we plot the $r_p$ (Figure \ref{fig:PearsonGPU}) and $r_s$ (Figure \ref{fig:SpearmanGPU}) with batch size one and four. The trend in variation of $r_p$ and $r_s$ with $\alpha$ is similar across both  $B$=1 and $B$=4. Also, there is negligible change in absolute values of $r_p$ and $r_s$ at $\alpha$=0.8. Thus, even though changing $B$ alters the energy/power consumption of different DNNs differently (Table \ref{tab:DataReuse}), the relationship between weighted arithmetic intensity and energy-efficiency of DNNs remains linear irrespective of $B$.

\begin{table}[htbp] \centering
  \caption{  Correlation ($r_p$ and $r_s$) of $AI_c$ and $DI$ with energy efficiency of 25 DNNs measured on P4000 and P100 GPUs with $B$=1 and $B$=4} 
   \label{tab:CorrelationComparison}
       \begin{tabular}{|C{1cm}|c|c|l|c|} \hline
{\bf Batch size} & {\bf GPU} & {\bf Correlation} & {\bf Metric} & {\bf CorrelCoeff.} \\ \hline
     \multirow{8}{*}{ $B$=1 } & \multirow{4}{*}{ P4000 } & \multirow{2}{*}{ PPMCC ($r_p$) } & $AI_c$ & 0.62\\
& & & $DI$ {\bf (ours)} & 0.88 \\ \cline{3-5}
& & \multirow{2}{*}{ SRCC ($r_s$)  } & $AI_c$ & 0.72 \\ 
& & & $DI$ {\bf (ours)} & 0.76 \\ \cline{2-5}
& \multirow{4}{*}{ P100 } & \multirow{2}{*}{ PPMCC ($r_p$) } & $AI_c$ & 0.63 \\
& & & $DI$ {\bf (ours)} & 0.87 \\ \cline{3-5}
& & \multirow{2}{*}{ SRCC ($r_s$)  } & $AI_c$ & 0.68 \\ 
& & & $DI$ {\bf (ours)} & 0.73 \\ \hline
\multirow{8}{*}{ $B$=4 } & \multirow{4}{*}{ P4000 } & \multirow{2}{*}{ PPMCC ($r_p$) } & $AI_c$ & 0.70 \\
& & & $DI$ {\bf (ours)} & 0.85 \\ \cline{3-5}
& & \multirow{2}{*}{ SRCC ($r_s$)  } & $AI_c$ & 0.72 \\ 
& & & $DI$ {\bf (ours)} & 0.79 \\ \cline{2-5}
& \multirow{4}{*}{ P100 } & \multirow{2}{*}{ PPMCC ($r_p$) } & $AI_c$ & 0.66 \\
& & & $DI$ {\bf (ours)} & 0.86 \\ \cline{3-5}
& & \multirow{2}{*}{ SRCC ($r_s$)  } & $AI_c$ & 0.59 \\ 
& & & $DI$ {\bf (ours)} & 0.73 \\ \hline
\end{tabular} 
\end{table}

Increasing $B$ increases the data level parallelism and arithmetic intensity \cite{mlsys2020_12}.
More precisely, it increases the weight reuse; however, the activation reuse remains constant with a change in $B$. The variation in data reuse in a DNN is either due to different layer types (Conv, FC, non-Conv) or layers' dimensions. Therefore, when activation reuse is very low due to the presence of a particular type of convolution, then energy-efficiency seldom improves with a higher $B$. For example, both DWConv and FC layers are bandwidth-bound because the former has very low activation reuse (and higher weight reuse), whereas the latter has low weight reuse (and higher activation reuse) \cite{park2018deep}. However, FC layers become compute-intensive at higher $B$ \cite{mlsys2020_12}; whereas energy-efficiency of the DWConv layer does not improve with increasing $B$. Thus, if our proposed model would have been sensitive to $B$, the importance of weight reuse would have reduced further, and the saturation points in Figure \ref{fig:PearsonGPU} would have shifted towards the right (i.e., $\alpha$ would exceed $0.8$). In reality, saturation points remain the same across both $B$=1 and $B$=4. {\em Evidently, our model remains valid for different values of  $B$}.

In Table \ref{tab:CorrelationComparison}, we compare $DI$ with $AI_c$ using $r_p$ and $r_s$. Evidently, compared to $AI_c$, $DI$ has a stronger correlation ($r_p$) with energy efficiency as measured for 25 DNNs on both P4000 and P100 GPUs. However, the difference between $r_s$ for $AI_c$ and $DI$ is quite small as compared to the difference between $r_p$ for $AI_c$ and $DI$ (Table \ref{tab:CorrelationComparison}). For example, the difference between $r_s$ is only 0.04, whereas that between $r_p$ is 0.26. This further strengthens our claim that there is a group of DNNs for which the conventional metric $AI_c$ does not have a linear relationship with energy-efficiency, but  $DI$ does have a linear relationship. Hence, $DI$ is a better indicator of the energy efficiency of DNNs. From Table \ref{tab:ConvType}, we conclude that {\em its the activation reuse which causes variation in arithmetic intensity, not the weight reuse}. In our model, we obtained $\alpha=0.80$, which indicates that activation reuse has a much higher impact on $DI$ than weight reuse. Evidently, our model confirms the intuition (Section \ref{subsec:WhyUnequal}) and highlights the importance of activation reuse.

\begin{table*} [htbp]
\caption{Comparison of data reuse in terms of weight ($\frac{M_c}{W}$) and activation reuse ($\frac{M_c}{A}$); conventional metric ($AI_c$) and {\bf proposed metric} ($DI$) for 25 DNNs. Also, the disparity factor ($d_f$), activation to parameter ratio ($\frac{A}{W}$), along with average power ($P_{avg}$ in watts) and inference time ($I_t$ in milliseconds) measured for $B$=1 and $B$=4 on P4000 and P100 GPUs.} 
\label{tab:DataReuse}
\centering
\resizebox{0.95\textwidth}{!}{
\begin{tabular}{|c|c|c|c|c|c|c|c||c|c|c|c|c|c|c|c|} \hline
\multirow{3}{*}{\bf Model Name} & \multirow{3}{*}{\bf Image size} & \multirow{3}{*}{ \textbf{$\frac{M_c}{W}$}} & \multirow{3}{*}{\textbf{$\frac{M_c}{A}$}} & \multirow{3}{*}{ \textbf{$AI_c$} } & \multirow{3}{*}{ \textbf{$DI$} } & \multirow{3}{*}{ $d_f$ } & \multirow{3}{*}{$\frac{A}{W}$} & \multicolumn{4}{c|}{\bf P100 } & \multicolumn{4}{c|}{\bf P4000 } \\ \cline{9-16}
& & & & & & & & \multicolumn{2}{c|}{ $B$=1 } & \multicolumn{2}{c|}{ $B$=4 } & \multicolumn{2}{c|}{ $B$=1 } & \multicolumn{2}{c|}{ $B$=4 } \\ \cline{9-16}
& & & & & & & & $I_t$ & $P_{avg}$ & $I_t$ & $P_{avg}$ & $I_t$ & $P_{avg}$ & $I_t$ & $P_{avg}$ \\ \hline
AlexNet\cite{NIPS2012_4824} & $224\times224$ & 11.85 & 361.50 & 11.48 & 72.89 & {\bf -535.16} & 0.03 & 2.21 & 37.0 & 2.92 & 50.8 & 2.40 & 60.5 & 3.77 & 68.6 \\
VGG-16\cite{Simonyan2014VeryDC} & $224\times224$ & 111.81 & 537.15 & 92.55 & 113.02 & -22.12 & 0.21 & 9.19 & 58.3 & 21.83 & 74.8 & 10.50 & 76.0 & 34.00 & 83.8 \\
NiN\cite{DBLP:journals/corr/LinCY13} & $224\times224$ & 146.05 & 291.34 & 97.28 & 65.57 & 32.60 & 0.50 & 2.88 & 35.0 & 3.69 & 59.0 & 2.30 & 59.0 & 4.45 & 73.0 \\
GoogLeNet\cite{Szegedy_2015_CVPR} & $224\times224$ & 227.14 & 158.05 & 93.20 & 42.97 & 53.90 & 1.44 & 10.64 & 20.5 & 12.50 & 40.8 & 8.36 & 44.2 & 11.50 & 64.2 \\
Inception-V2\cite{Szegedy_2016_CVPR} & $231\times231$ & 196.43 & 122.22 & 75.34 & 34.27 & 54.52 & 1.61 & 18.54 & 17.3 & 21.40 & 31.7 & 14.50 & 41.6 & 19.70 & 57.0 \\
Inception-V3\cite{Szegedy_2016_CVPR} & $299\times299$ & 240.34 & 138.40 & 87.82 & 39.70 & 54.80 & 1.74 & 29.40 & 21.5 & 39.80 & 38.4 & 24.91 & 48.1 & 38.12 & 65.2 \\
Inception-V4\cite{Szegedy2017Inceptionv4IA} & $299\times299$ & 287.49 & 169.10 & 106.47 & 48.19 & 54.74 & 1.70 & 55.83 & 24.0 & 71.00 & 41.3 & 41.45 & 51.2 & 72.50 & 67.0 \\
ResNet-50\cite{He_2016_CVPR} & $224\times224$ & 151.41 & 82.83 & 53.54 & 24.14 & 54.92 & 1.83 & 15.47 & 27.3 & 23.60 & 43.5 & 14.56 & 53.1 & 29.20 & 66.8 \\
ResNet-101\cite{He_2016_CVPR} & $224\times224$ & 170.37 & 107.51 & 65.91 & 30.02 & 54.46 & 1.58 & 32.06 & 26.0 & 42.83 & 42.0 & 26.50 & 52.6 & 48.70 & 66.0 \\
ResNet-152\cite{He_2016_CVPR} & $224\times224$ & 187.74 & 112.88 & 70.49 & 31.96 & 54.66 & 1.66 & 46.42 & 25.5 & 62.10 & 41.4 & 38.00 & 52.2 & 69.60 & 66.5 \\
ResNet101-V2\cite{He2016IdentityMI} & $224\times224$ & 175.57 & 112.20 & 68.45 & 31.22 & 54.40 & 1.56 & 32.26 & 26.5 & 43.00 & 42.5 & 26.00 & 53.0 & 49.44 & 56.3 \\
ResNet152-V2\cite{He2016IdentityMI} & $224\times224$ & 191.56 & 116.23 & 72.34 & 32.82 & 54.62 & 1.65 & 44.13 & 26.9 & 63.13 & 42.5 & 38.38 & 53.5 & 71.30 & 55.8 \\
Inception-ResNet-V2\cite{Szegedy2017Inceptionv4IA} & $299\times299$ & 235.85 & 139.04 & 87.47 & 39.60 & 54.73 & 1.70 & 71.10 & 24.7 & 92.48 & 44.0 & 56.30 & 50.3 & 89.64 & 70.4 \\
ResNext50-32x4d\cite{Xie2017AggregatedRT} & $224\times224$ & 190.97 & 71.59 & 52.07 & 23.87 & 54.16 & 2.67 & 23.50 & 25.8 & 34.27 & 41.2 & 22.24 & 50.0 & 43.25 & 64.5 \\
ResNext101-32x4d\cite{Xie2017AggregatedRT} & $224\times224$ & 205.52 & 89.00 & 62.11 & 28.08 & 54.79 & 2.31 & 53.56 & 22.8 & 68.00 & 37.1 & 43.84 & 47.7 & 74.80 & 63.0 \\
DenseNet-121\cite{8099726} & $224\times224$ & 385.96 & 44.01 & 39.50 & 28.10 & 28.87 & 8.77 & 35.82 & 16.6 & 47.32 & 32.4 & 30.00 & 42.0 & 52.02 & 58.4 \\
DenseNet-169\cite{8099726} & $224\times224$ & 262.90 & 43.08 & 37.01 & 21.76 & 41.21 & 6.10 & 49.70 & 16.3 & 65.30 & 31.0 & 42.00 & 41.8 & 69.45 & 56.5 \\
SqueezeNet-V1.0\cite{Iandola2016SqueezeNetAA} & $224\times224$ & 678.08 & 68.91 & 62.55 & 47.69 & 23.77 & 9.84 & 3.69 & 22.2 & 5.45 & 43.1 & 3.40 & 46.1 & 6.80 & 65.5 \\
SqueezeNet-V1.1\cite{Iandola2016SqueezeNetAA} & $224\times224$ & 281.57 & 48.49 & 41.37 & 23.78 & 42.52 & 5.81 & 3.42 & 17.3 & 4.23 & 31.4 & 2.96 & 41.1 & 4.72 & 56.6 \\
1.0-SqNxt-23\cite{Gholami2018SqueezeNextHN} & $224\times224$ & 380.50 & 15.32 & 14.73 & 22.09 & -50.00 & 24.84 & 22.35 & 9.4 & 24.22 & 15.7 & 18.41 & 33.1 & 22.34 & 42.0 \\
1.0-SqNxt-23v5\cite{Gholami2018SqueezeNextHN} & $224\times224$ & 242.04 & 16.01 & 15.02 & 15.30 & -1.91 & 15.12 & 24.56 & 8.3 & 25.88 & 14.0 & 18.00 & 32.4 & 20.85 & 40.0 \\
1.0-G-SqNxt-23\cite{Gholami2018SqueezeNextHN} & $224\times224$ & 406.35 & 12.39 & 12.02 & 22.80 & -89.61 & 32.80 & 24.27 & 9.3 & 25.81 & 15.3 & 18.80 & 33.4 & 22.22 & 42.3 \\
MobileNet-V1\cite{Howard2017MobileNetsEC} & $224\times224$ & 135.65 & 28.24 & 23.37 & 12.43 & 46.82 & 4.80 & 27.71 & 9.5 & 93.97 & 9.6 & 19.10 & 35.0 & 64.40 & 35.4 \\
MobileNet-V2\cite{8578572} & $224\times224$ & 124.80 & 12.36 & 11.24 & 8.71 & 22.52 & 10.10 & 42.90 & 9.5 & 140.11 & 10.4 & 30.30 & 34.7 & 98.23 & 35.3 \\
XceptionNet\cite{Chollet_2017_CVPR} & $299\times299$ & 366.58 & 84.81 & 68.87 & 35.29 & 48.76 & 4.32 & 124.92 & 11.4 & 468.16 & 11.5 & 87.20 & 38.3 & 324.36 & 38.5 \\ \hline
\end{tabular}}
\end{table*}

\subsection{Salient features of our model}\label{sec:SalientFeatures}
We first discuss how modeling of activation and weight reuse captures all types of reuses across different layers in DNNs. 
Based on the computation and data reuse patterns, layers in DNNs can be broadly categorized as Conv, non-Conv, and FC layers. In Conv layers, there is feature map reuse, filter reuse, and filter weight reuse \cite{Yang2017AMT}, whereas in FC layers, except filter-weight reuse,  all these types of reuses are possible. In other words, all the possible data reuse in both Conv and FC layers can be expressed as weight and activation reuse. However, in non-Conv layers, only ifmap and ofmap activations are processed, and their reuse can be described as activation reuse. 

As shown in Table \ref{tab:ConvType}, the degree of data reuse in different types of convolutions is different due to the activation reuse, whereas weight reuse remains constant across all types of convolutions. Hence, energy efficiency of MACs for different types of convolutions can be expressed in terms of activation reuse.  Since non-Conv layers do not have learn-able parameters and process only feature maps, their energy metrics are governed by activation reuse. However, in  FC layers,  $M_c \approx W$ and $A \ll W$, thus,  arithmetic intensity and weight reuse are approximately equal to 1. Hence, the energy efficiency of MACs in FC layers is governed by weight reuse. 

In summary, except FC layer, all the layers' energy efficiency can be expressed in terms of activation reuse. The higher value of $ \alpha $ implies higher importance of activation reuse. Hence, FC layers have a lower impact on the energy efficiency of DNNs. In fact, deeper networks such as Inception-V4, Inception-ResNet-V2 have hundreds of Conv and non-Conv layers, but very few FC layers and some networks such as NiN have no FC layers at all.

We now discuss how our model addresses the shortcomings of $AI_c$. As discussed in Section \ref{subsec:CumulativeAI}, $AI_c$ predicts AlexNet as memory-bound and MobileNet-V1 as compute-bound (shown in roofline models in Figure \ref{fig:RooflineP4000} and \ref{fig:RooflineP100}) but the energy-efficiency of AlexNet is quite high  and that of the MobileNet-V1 is quite low (Figure \ref{fig:RooflineCorrelationP4000} and \ref{fig:RooflineCorrelationP100}). This is counter-intuitive because operations (MACs) of memory-bound workload are energy-inefficient due to the higher number of memory-accesses. The reason for these irregularities are better explained by our model. As shown in Table \ref{tab:DataReuse}, AlexNet has very low $\frac{M_c}{W}$ but significantly high $\frac{M_c}{A}$, whereas MobileNet-V1 has high $\frac{M_c}{W}$ but significantly low $\frac{M_c}{A}$.  
By virtue of giving higher importance to $\frac{M_c}{A}$, our model is able to accurately predict AlexNet as compute-bound  and MobileNet-V1 as memory-bound (Figure \ref{fig:RooflineAIwP4000}  and \ref{fig:RooflineAIwP100})

\subsection{When and why does $AI_c$ fail? } \label{sec:AIfailure}

We define relative disparity ($d_f$) between $AI_c$ and $DI$ as
\begin{equation}
 \label{eqn:RelativeDisparity}
 d_f = \Bigg(\frac{AI_c - DI}{AI_c}\Bigg)\times 100 = 75 - 6.25\times \bigg[ \frac{A}{W} + 3\times \frac{W}{A} \bigg]
\end{equation}

 Equation  \ref{eqn:RelativeDisparity} shows that,  $\frac{A}{W}$  has less impact on relative disparity ($d_f$) than $\frac{W}{A}$. Since  $\frac{A}{W}$ is same as weight reuse to activation reuse ratio, Eq. \ref{eqn:RelativeDisparity} implies that weight reuse has less impact on $d_f$. Table \ref{tab:DataReuse} shows weight reuse ($\frac{M_c}{W}$),  activation reuse ($\frac{M_c}{A}$), $AI_c$ and $DI$, activation to parameter ratio ($\frac{A}{W}$) and $d_f$ value  for 25 state-of-the-art DNNs. For gaining more insights, we study three cases which are shown in Table \ref{tab:AIfailure}.

\begin{table}[htbp]
\caption{Disparity between $AI_c$ and $DI$ for  different cases}
\label{tab:AIfailure}
\centering
\resizebox{0.48\textwidth}{!}{
\begin{tabular}{ |c |c |c |c |} 
 \hline 
 & Case 1: $A \ll W$ & Case 2: $A \approx W$ & Case 3: $A \gg W$ \\ \hline
  $M_c$/$A$      & larger   & comparable & smaller\\
  $M_c$/$W$      & smaller  & comparable &larger \\
  $AI_c$       &  $\approx  M_c$/$W $       & $\approx 0.5\times M_c$/$A$ & $\approx M_c$/$A$\\
  $DI$      & $\approx 0.2\times M_c$/$A$         & $\approx 0.25\times M_c$/$A$ &$\approx 0.06\times M_c$/$W$\\
  $d_f$  & $\approx 75-18.75\times \frac{W}{A}$ & $\approx 50$ & $\approx 75-6.25\times \frac{A}{W}$ \\
 \hline
\end{tabular} }
\end{table}

As shown in Table \ref{tab:AIfailure}, in {\bf case 1},  activation reuse ($\frac{M_c}{A}$) dominates total data reuse ($\frac{M_c}{W}$ + $\frac{M_c}{A}$), but $AI_c$ is nearly equal to weight reuse ($\frac{M_c}{W}$). This leads to huge disparity between $AI_c$ and $DI$. For example, AlexNet has $30\times$ higher activation reuse than weight reuse and hence, its relative disparity is highest among all the 25 DNNs (Table \ref{tab:DataReuse}). In {\bf case 3}, weight reuse dominates the total data reuse, however, disparity is noticeable but not as large as in case 1 because weight reuse has less impact compared to activation reuse (refer Eq. \ref{eqn:WeightedAI}). In {\bf case 2}, the $d_f$ is lower and $AI_c$ would be able captures the dynamics of data reuse in DNNs. In summary, when either $A \approx W$, for example in variants of InceptionNet, ResNet and ResNext (Table \ref{tab:DataReuse}), or when $A$ is significantly  higher than $W$, for example MobileNet-V2 and variants of SqueezeNext (Table \ref{tab:DataReuse}), $AI_c$ would be able to capture the data reuse patterns in DNNs. {\em However, $AI_c$ fails to estimate the data reuse when the  $\frac{A}{W}$ ratio is very low (e.g., AlexNet) and also, when $\frac{A}{W}$ ratio is moderately high (e.g., MobileNet-V1).}

\section{Generality and Use Cases of $DI$} \label{sec:GeneralityTesting}

\subsection{Generality of Proposed Model}
For the generality test, we use confidence intervals for the population correlation coefficient ($\rho$), which measures the linear correlation between two variables over the entire population. Note that the sample (Pearson) correlation coefficient ($r_p$) is a measure of the correlation between the two variables over a sample ($\phi$) taken randomly from the population. We perform the following steps to compute the confidence intervals for $\rho$ at a given $r_p$ and $\phi$.

\textbf{Step 1:} 
``Central limit theorem'' can be applied when the data follow normal distribution or the sample size is large. In our experiment, $\phi$ is 25 which is large enough (refer Eq. \ref{Eqn:MinNoOfDNN}) to apply ``central limit theorem''. To get normal distribution, we transform $r_p$ using ``Fisher's $Z$ transform'' as $Z_r = \frac{1}{2}\times \log_e\big(\frac{1+r_p}{1-r_P}\big)$ \cite{CorrelationAndRegression}. Then, we calculate standard error ($S_e$) which is approximated as $\frac{1}{\sqrt{\phi - 3}}$ \cite{CorrelationAndRegression}.

\textbf{Step 2}: For 95\% confidence, the upper limit ($U_{r_p}$) and lower limit ($L_{r_p}$) of confidence intervals  are $U_{r_p}$ = $Z_{r_p} + (1.96\times S_e)$ and $L_{r_p}$ = $Z_{r_p} -  (1.96\times S_e)$ respectively. Similarly, for 99\% confidence, $U_{r_p}$ = $Z_{r_p} + (2.58\times S_e)$ and $L_{r_p}$ = $Z_r - (2.58\times S_e)$ respectively \cite{CorrelationAndRegression}. Note that these confidence intervals are corresponding to $Z_{r_p}$. 

\textbf{Step 3:} We compute inverse Fisher transform \cite{CorrelationAndRegression} to calculate the upper ($U$) and lower ($L$) limit of confidence intervals corresponding to $r_p$,   which are $U$ = $\frac{e^{2\times {U_{r_P}}} - 1}{e^{2\times {U_{r_p}}} + 1} $ and $L$ = $\frac{e^{2\times {L_{r_p}}} - 1}{e^{2\times {L_{r_p}}} + 1} $.

\textbf{Observations:} Table \ref{tab:ConfidenceInterval} shows the confidence interval for both 95\% and 99\% confidence. For a better approximation of $\rho$ using $r_p$, window size ($\Delta$) for a confidence interval, should be as narrow as possible. Smaller window size implies a lesser deviation in $\rho$ and ensures that the same correlation would hold in other sets of samples taken from a large population. For $AI_c$ on P4000 GPU, at 99\% confidence, $L$=0.31 and $\Delta$=0.58. On P100 GPU, these values are $L$= 0.27 and $\Delta$=0.63. This shows that, $AI_c$ can have very poor correlation with energy efficiency in some cases, e.g., for AlexNet, relative disparity ($d_f$) is very high  (Table \ref{tab:DataReuse}). By comparison, with $DI$ at 99\% confidence, $L$=0.61 and $\Delta$=0.34 on P4000 GPU, whereas $L$=0.63 and $\Delta$=0.32 on P100 GPU. Thus, it can be said with 99\% confidence  that $\rho$ lies between 0.61 to 0.95 on P4000 and between 0.63 to 0.95 on P100 GPU.  Clearly, {\em our model exhibits better positive correlation with energy efficiency in any population of DNNs on both GPUs.}

\begin{table} [htbp] \centering
  \caption{Confidence intervals for population correlation ($\rho$)} \vspace{-0.2cm}
   \label{tab:ConfidenceInterval}
  
  \begin{tabular}{ |c|c| ccc| ccc| }
    \hline
    \multirow{2}{*}{ } & { } &
      \multicolumn{3}{c|}{95\% confidence} &
      \multicolumn{3}{c|}{99\% confidence } \\
      
     { GPU}  & {\textbf{Metric} } & {$L$} & {$U$} & { $\Delta$ } & {$L$} & {$U$} & { $\Delta$ } \\
      \hline
   \multirow{2}{*}{ P4000} &$AI_c$  & 0.42 & 0.86 & 0.44 & 0.31 & 0.89 & 0.58 \\
    &$DI$ (\textbf{Ours}) & 0.68 & 0.93 & \textbf{0.25} & 0.61 & 0.95 & \textbf{0.34} \\ 
    \hline
   \multirow{2}{*}{ P100} &$AI_c$ &0.36 & 0.84 & 0.48 & 0.24 & 0.87 & 0.63 \\
    &$DI$ (\textbf{Ours}) & 0.70 & 0.94 & \textbf{0.24} & 0.63 & 0.95 & \textbf{0.32} \\ 
    \hline
  \end{tabular}
\end{table}

\textbf{Minimum $\phi$ required for   generality test:} Unlike $\Delta$, window size corresponding to $Z_{r_p}$, i.e. $\Delta_r$ = $U_{r_p}$ - $L_{r_p}$, is independent of $r_p$. We find minimum sample size ($\phi$) such that $\Delta_{r_p} \leqslant 1$ at 95\% confidence.

\begin{equation} \label{Eqn:MinNoOfDNN} 
\Delta_{r_p} \leqslant 1  \implies 
2\times 1.96\times \frac{1}{\sqrt{\phi - 3}}  \leqslant 1  \implies \phi \geqslant 18.37
\end{equation} 
%\vspace{-0.3cm}

The minimum number of DNNs required for the generality test is 19. We take $\phi$ as 25 for which $\Delta_{r_p}$  = 0.836.

\subsection{Use Cases for Proposed Model}
The design process of a DNN (e.g., pruning, quantization) requires massive human efforts to fine-tune the design hyper-parameters \cite{2018_ECCV_HeAMC}. In DNN pruning techniques, deciding the compression ratio for each layer is a daunting task, especially for a deeper network such as ResNet-152 \cite {He_2016_CVPR} and Inception-V4 \cite {Szegedy_2016_CVPR}. To save these human efforts, there is a growing trend for automating the design of machine learning models, which is termed as AutoML. For example, He et al. \cite{2018_ECCV_HeAMC} automate the task of pruning using reinforcement learning. Similarly, Wang et al. \cite{2019_CVPR_WangHAQ} automate the process of layer-wise quantization using reinforcement learning. Furthermore, there is an increasing trend in automating the design of compact and high-performance DNNs \cite{2019_ICML_TanEfficientNet}.

The works mentioned above on automating the design process optimize the DNN architecture to reduce the number of MAC operations. Since these metrics are only a proxy for energy consumption, optimizing them does not necessarily optimize the energy efficiency of DNNs. To design energy-efficient DNNs, a design-time metric that is a better representative of energy-efficiency of DNNs is required. Such a metric can be used in the objective function of AutoML tasks. Since data movements primarily drive energy consumption, data reuse can be used as the approximation for memory accesses ($\frac{1}{DI}$). Hence, instead of using \#MACs for the energy-efficient design of DNNs, the following metric can be used in the objective functions in the AutoML tasks.
  \begin{align} \label{eqn:OptimizedMetric}
 \text{Optimized metric} \propto M_c\times \Big(\frac{1}{DI}\Big)^{k}\; s.t.\; k\in(0,1)
 \end{align}
 
Since the cost of memory access is orders of magnitude higher than arithmetic operation, $k$ is used to normalize the memory cost with respect to computational cost in Eq. \ref{eqn:OptimizedMetric}. Consequently, the above optimized metric balances the number of computations and the number of memory access to optimize the network's energy efficiency. Evidently, $DI$ will be a  valuable tool for DNN designers.

\section{Validity of Proposed Model} \label{sec:ApplicabilityOfModel}

Our model estimates the data reuse available in DNNs with the assumption that underlying platforms have sufficient compute/memory resources to exploit the available data reuse in DNNs.  However, different hardware platforms are optimized for contrasting design goals and have dissimilar memory-hierarchy with a non-identical number of layers and capacity. We now discuss whether our model applies to general-purpose hardware such as GPU and CPU, or do we need to re-calibrate the value of $\alpha$ on them? We also discuss the effect of memory-hierarchy on our model.

{\bf GPU:} Both P100 and P4000 GPUs have substantially different memory and compute capability, which is also manifested by different CMR values (Table \ref{tab:GPUsFeatures}). Hence, the memory-access pattern and their cost would vary significantly for both the GPU. In Figure \ref{fig:ExperimentalGraphs}(a), it is shown that EPP values of DNNs are higher on P4000 GPU compared to that on P100 GPU. Our extensive experiments validate the proposed model and hence do not require re-calibration of $ \alpha $ for both the GPUs, even though they have different compute and memory resources. This indicates the applicability of our model to different GPUs regardless of their memory-hierarchy and CMR values. However, one may need to re-calibrate the $ \alpha $ to use $ DI $ as a representative of the energy efficiency of other DNN models that are not used in our experiments. 

\begin{figure}[htbp]
\centering
\includegraphics[scale=0.45]{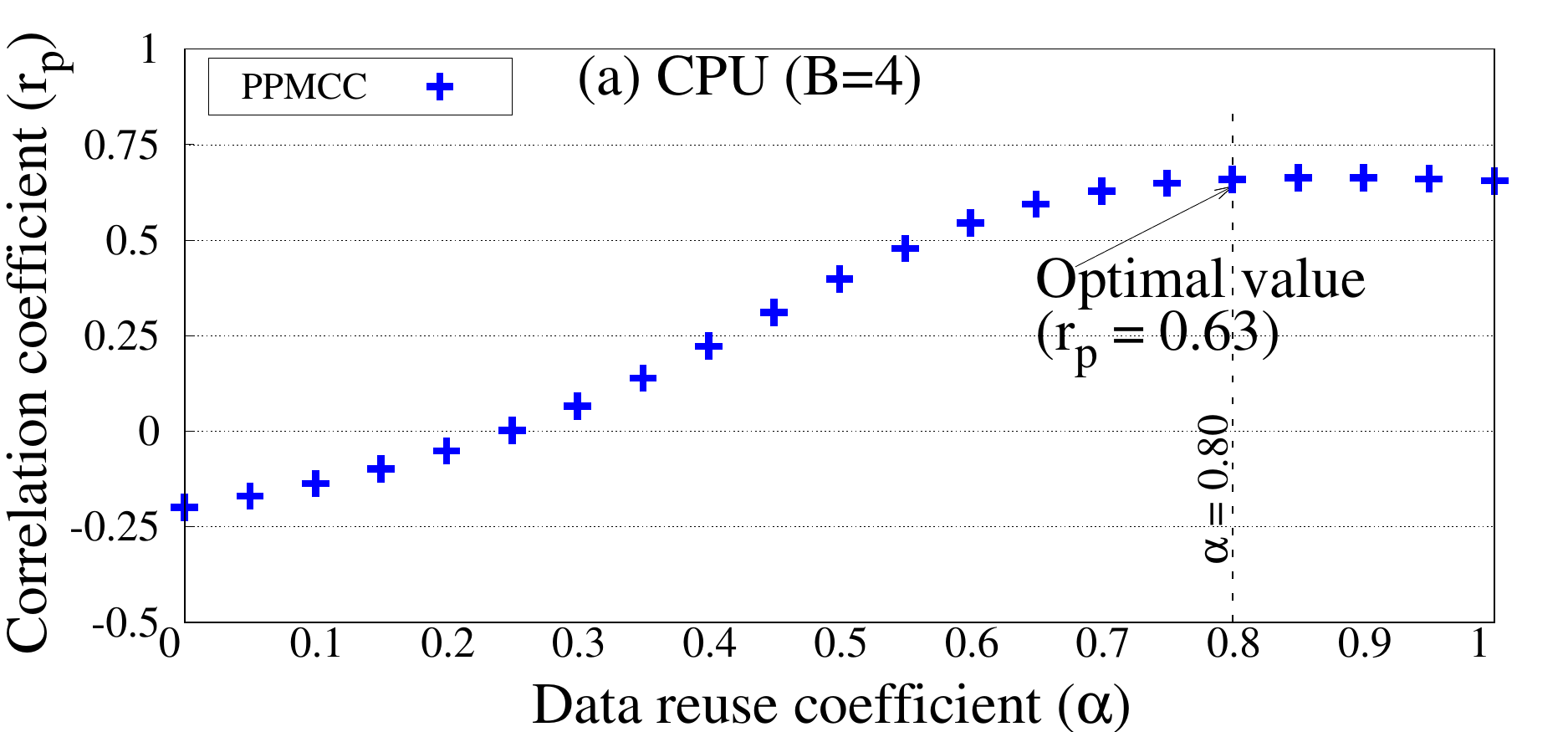}
\includegraphics[scale=0.45]{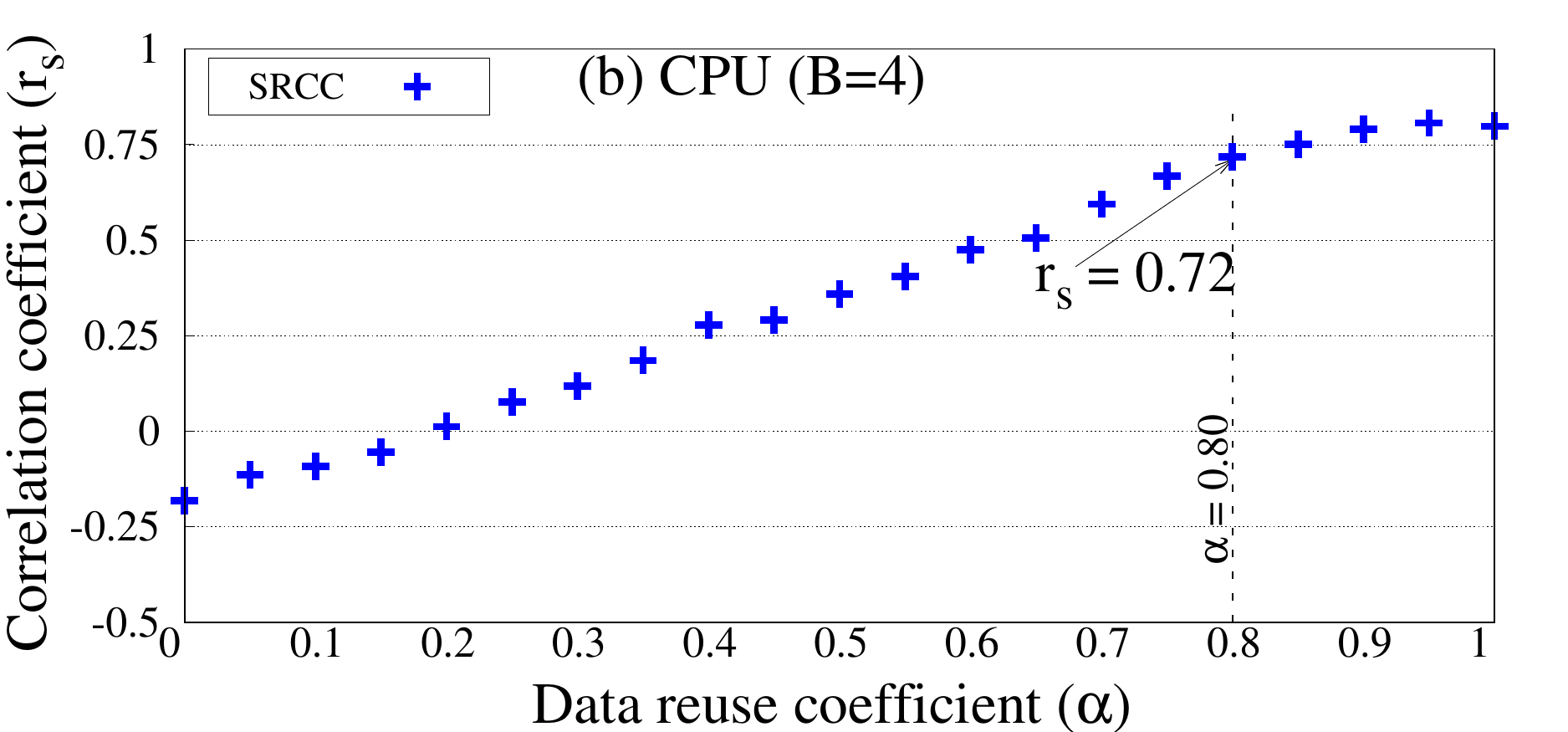}
\caption{PPMCC (a) and SRCC (b) calculated between weighted arithmetic intensity at different values of $\alpha$, and energy efficiency measured for 25 DNNs on CPU with $B$=4}
\label{fig:CorrelationComparsionCPU}
\end{figure}

{\bf CPU: }  CPUs have hardware managed cache hierarchy along with sophisticated techniques for cache miss management and cache coherence. Also, CPUs have a higher amount of off-chip memory than GPUs, which can accommodate a larger model with larger batch size. Furthermore, CPUs have a much lower amount of parallelism than GPUs. 
We investigate whether the above-mentioned differences affect our model. We perform our experiments on ``Intel(R) Xeon(R) CPU E5-1650 v4 @ 3.60GHz" which has 12 cores, 64KB L1 cache, 256KB L2 cache, 15MB L3 cache, and 32GB primary memory. The correlation results are shown in Figure \ref{fig:CorrelationComparsionCPU}. Similar to the results on P100 and P4000 GPU, both $r_p$ and $r_s$ increase with rising value of $\alpha$ and $r_p$ saturates at 0.8. This substantiates the higher importance of activation reuse for estimating the available data reuse. Therefore, the correlation trends would be similar (PPMCC first increases with $\alpha$ and later saturates at higher $\alpha$) even for the DNNs not used in our experiments. However, the value of  $\alpha$ may need to be re-calibrated for the precise use of  $DI$ as a representative of the energy-efficiency of other DNN models that are not used in our experiments, on the CPU.

\begin{figure}
\includegraphics[scale=0.45]{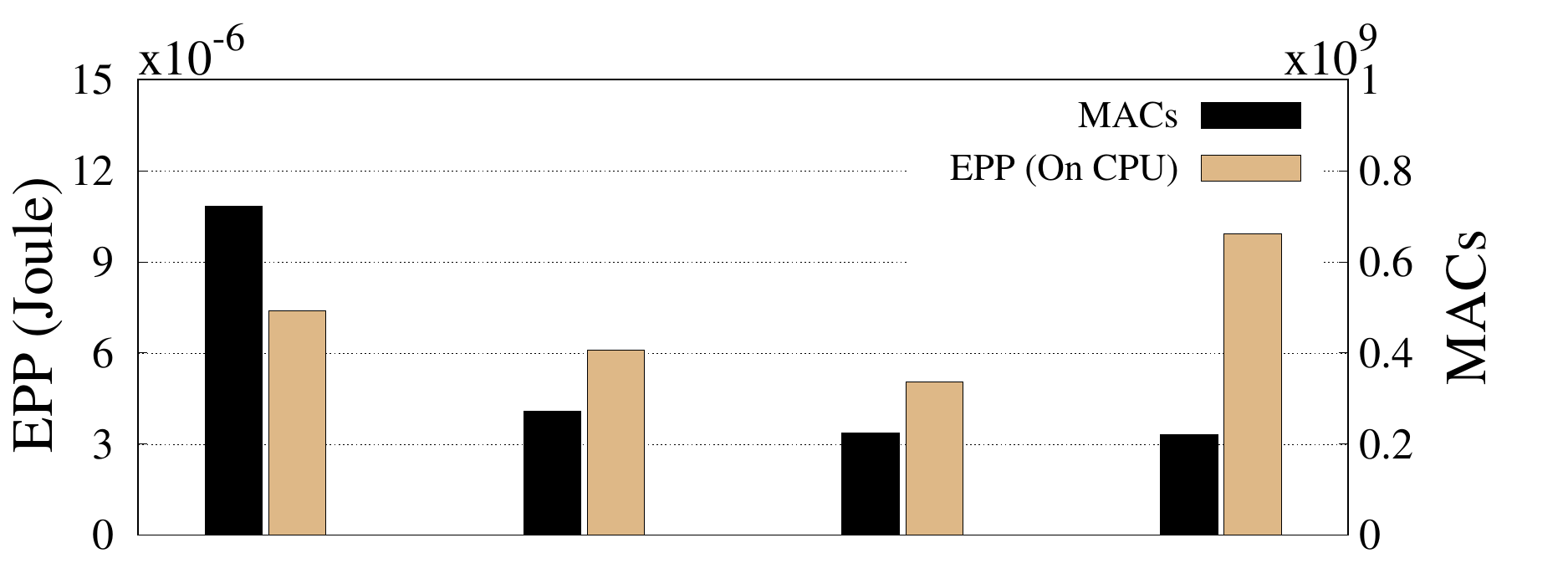} 
\includegraphics[scale=0.45]{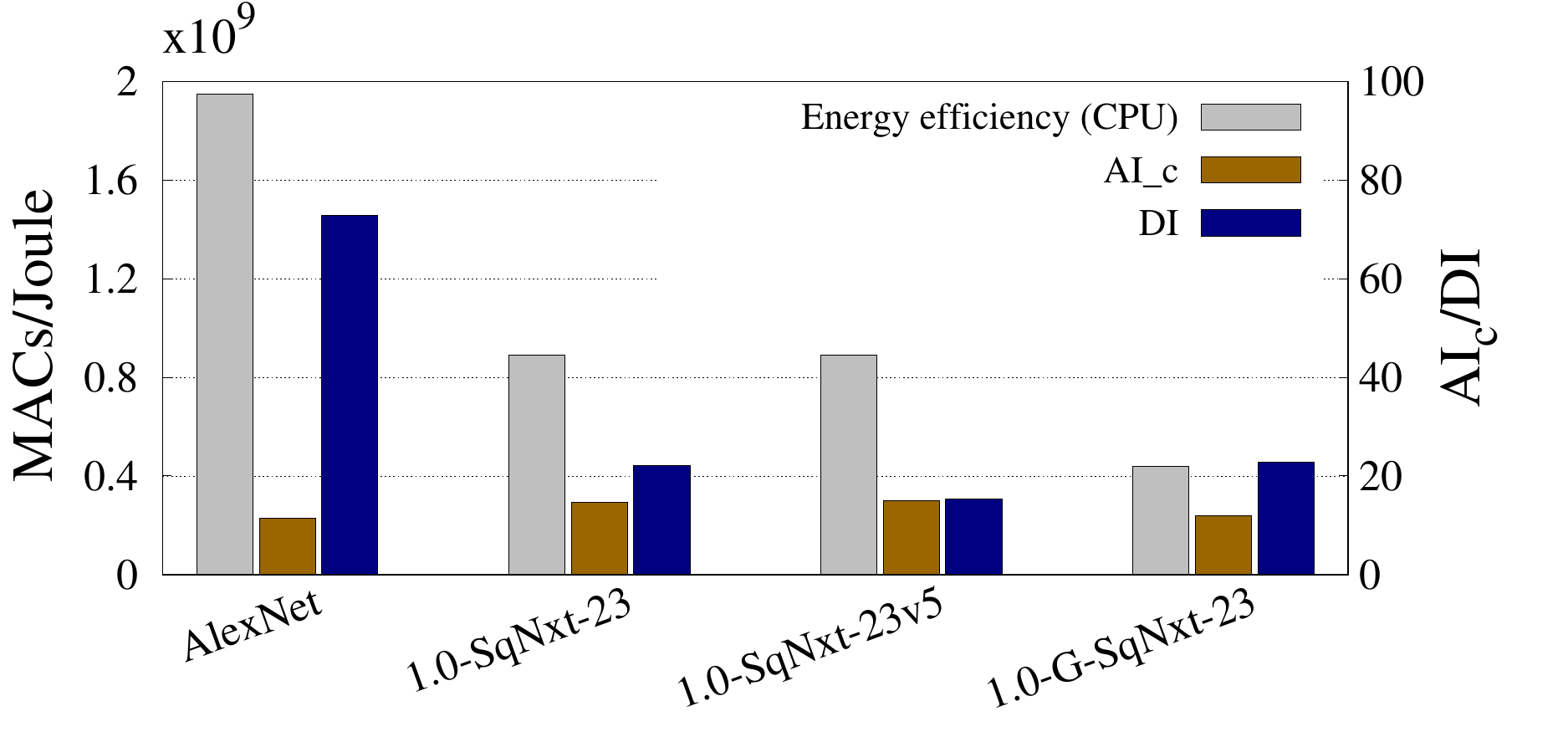}
\caption{Energy consumption (EPP) and energy-efficiency of AlexNet and SqueezeNext variants measured on CPU.} \label{fig:CPUEnergy}
\end{figure}

For a better comparison of our model with $AI$ on CPU, we select AlexNet and three SqueezeNext variants, which have a very high  disparity between $A$  and $W$  (Table \ref{tab:DataReuse}). The EPP, energy efficiency, $AI_c$, and $ DI $ are shown in Figure \ref{fig:CPUEnergy}. Also, the calculated $r_p$ and $r_s$ with $AI_c$ and $DI$ are shown in Table \ref{tab:CPUCorrelation}. The negative correlations ($r_p$ and $r_s$) with  $AI$ show the ineffectiveness of $AI$ as a representative of energy-efficiency on CPU, whereas, high positive correlations with $DI$ indicate $DI$ as an appropriate representative of energy-efficiency on CPU.

\begin{table}[htbp]
  \caption{Correlation ($r_p$, $r_s$) of $AI_c$ and $DI$ with energy efficiency of AlexNet and SqueezeNext variants on CPU} 
   \label{tab:CPUCorrelation}
   \resizebox{0.48\textwidth}{!}{
  \begin{tabular}{ |c|l|c|c| }
    \hline
    \multirow{2}{*}{ } 
      CPU  & $X$  & $Y$ & \textbf{correlation ($X$,$Y$)}  \\
      \hline
   \multirow{2}{*}{ PPMCC ($r_p$)} &$AI_c$  & Energy efficiency & -0.39 \\
    &$DI$ (\textbf{Ours})& Energy efficiency  & \textbf{0.91} \\ \hline
   \multirow{2}{*}{ SRCC ($r_s$)} &$AI_c$ &Energy efficiency & -0.37  \\
    &$DI$(\textbf{Ours})&Energy efficiency & \textbf{0.76} \\ 
    \hline
  \end{tabular}}
\end{table}

\section{Related Work} \label{sec:RelatedWork}

Yang et al. \cite{Yang2017AMT} proposed an energy estimation tool which takes the layer's dimensions and sparsity in DNN as inputs and estimates the energy consumption of a DNN. They validated their energy model on a systolic array-based processor (Eyeriss\cite{2017_Chen_JSSC}) with very few DNNs (AlexNet, VGG-16, GoogLeNet, and SqueezeNet). Chen et al. \cite{chen2019eyeriss} employed a model named ``Eyexam'', which takes design decisions of both the DNN model and systolic accelerator as input and predicts their effects on the performance. Kwon et al. \cite{2019_MICRO_Kwon} proposed an analytical cost model named ``MAESTRO'', which takes DNN's layer dimensions and the employed dataflow as inputs and provides detailed performance analysis of DNNs on systolic accelerators. Cai et al. \cite{cai2018proxylessnas} build a model that predicts the latency for latency-aware DNN design using an architecture search. Li et al. \cite{li2016evaluating} performed a detailed study on the energy-efficiency of DNNs on CPU and GPU and provides insights for the energy-aware design of DNNs. The works mentioned above demonstrate their effectiveness for very few DNNs, far from comprehensive, on a limited set of hardware platforms. They do not include state-of-the-art design methodologies, such as dense connections in DenseNets, where the number of activations varies at runtime. Hence, their models' applicability is quite limited and cannot be generalized over a wide range of DNNs. Nevertheless, Binaco et al. \cite{bianco2018benchmark} have done extensive experimentation and presented the detailed comparison of the inference time of the SOTA DNNs on a multitude of GPUs. However,  due to the missing energy comparison and the lack of an appropriate metric and/or mathematical model that can predict the inference time and/or energy efficiency of DNNs at the design time, the insights for data reuse in the representative DNNs is missing.

\section{Conclusion and future work} \label{sec:Conclusion}
We show that the conventional metric ($AI$) does not always accurately estimate the degree of data reuse in DNNs. We propose a novel model that decouples the weight and activation reuse and accounts for their unequal importance. We show that our model applies to a diverse set of DNNs and is a better representative of energy efficiency in DNNs.  In future work, we will evaluate our model on other accelerators to show its applicability over a broad range of hardware platforms.

{
\scriptsize
\linespread{0.97}
\bibliographystyle{IEEEtran1}
\bibliography{Ref}
}

\vspace{-1.5cm}
 \begin{IEEEbiography}[{\includegraphics[width=1in,height=1.05in,clip,keepaspectratio]{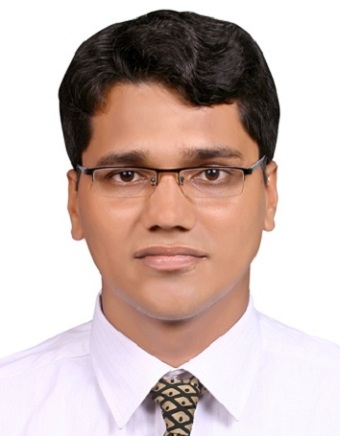}}]{Nandan Kumar Jha}
is an M.Tech. (Research Assistant) student in the CSE  at IIT, Hyderabad, India. He received the B.Tech. Degree in ECE from NIT, Surat, India, in 2013. He has worked as an Electrical Design Engineer in Seagate Technology, India and as a Project Research Assistant at IIT, Bombay. He research interests are  deep learning and computer architecture. 
\end{IEEEbiography}
\vskip -3\baselineskip plus -1fil
\begin{IEEEbiography}[{\includegraphics[width=1in,height=1.05in,clip,keepaspectratio]{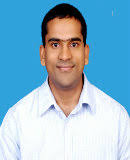}}]{Sparsh Mittal}
received the B.Tech. degree in ECE from IIT, Roorkee, India and the Ph.D. degree in computer engineering from Iowa State University, USA. He has worked as a Post-Doctoral Research Associate at Oak Ridge National Lab, USA and as an assistant professor at IIT Hyderabad, India. He is currently working as an assistant professor at IIT Roorkee. Sparsh has published nearly 90 papers in top conferences and journals. 
 
\end{IEEEbiography}

\end{document}